\documentclass[a4paper,fleqn]{cas-dc}

\usepackage[numbers]{natbib}
\usepackage{amsthm,amsfonts,amsmath,amssymb}
\usepackage{tabularx}  % For automatic column width adjustment
\usepackage[ruled,vlined]{algorithm2e}
\usepackage{pifont}
\usepackage{subfigure} 
\usepackage{graphicx}
\usepackage{psfrag}
\usepackage{pstool}

\usepackage{xcolor}
\usepackage{array}
\usepackage{units}
\usepackage{mdframed}
\usepackage{multicol}
\usepackage{colortbl}
\usepackage{setspace}
\usepackage{textcomp}
\usepackage{multirow}
\usepackage{enumitem}
\usepackage{booktabs} 
\usepackage{steinmetz}
\usepackage{tabularray}
\makeatletter
\newcommand{\mypm}{\mathbin{\mathpalette\@mypm\relax}}
\newcommand{\@mypm}[2]{\ooalign{%
  \raisebox{.1\height}{$#1+$}\cr
  \smash{\raisebox{-.6\height}{$#1-$}}\cr}}
\makeatother
\usepackage[ruled,vlined]{algorithm2e}

\def\tsc#1{\csdef{#1}{\textsc{\lowercase{#1}}\xspace}}
\tsc{WGM}
\tsc{QE}
\tsc{EP}
\tsc{PMS}
\tsc{BEC}
\tsc{DE}
\begin{document}
\let\WriteBookmarks\relax
\def\floatpagepagefraction{1}
\def\textpagefraction{.001}

%Short title
\shorttitle{\textcolor{blue}{}}

% Short author
\shortauthors{Shuyi Gao et~al.}

% Main title of the paper
\title [mode = title]{
Topology-Aware Graph Reinforcement Learning for Energy Storage Systems Optimal Dispatch in Distribution Networks}                  
% Title footnote mark
% eg: \tnotemark[1]
\tnotemark[1]

% Title footnote 1.
% eg: \tnotetext[1]{Title footnote text}
% \tnotetext[<tnote number>]{<tnote text>} 
% \tnotetext[1]{\textcolor{blue}{funding}}

% First author
%
% Options: Use if required
% eg: \author[1,3]{Author Name}[type=editor,
%       style=chinese,
%       auid=000,
%       bioid=1,
%       prefix=Sir,
%       orcid=0000-0000-0000-0000,
%       facebook=<facebook id>,
%       twitter=<twitter id>,
%       linkedin=<linkedin id>,
%       gplus=<gplus id>]
\author[1]{Shuyi Gao}
% [type=editor,
%                         style=chinese,auid=000,
% %                        bioid=1,
% %                        orcid=0000-0001-7511-2910
% ]

\credit{Conceptualization, Methodology, Software, Validation, Writing - original draft}

% Address/affiliation
\affiliation[1]{organization={Department of Electrical Sustainable Energy, Delft University of Technology},
    addressline={Mekelweg 4}, 
    city={Delft},
    postcode={2628 CD}, 
    country={The Netherlands}}

% \author[1]{Shengren Hou}
% % [type=editor,
% %                         style=chinese,auid=000,
% % %                        bioid=1,
% % %                        orcid=0000-0001-7511-2910
% % ]
% \credit{Writing - review \& editing}

% Second author

\author[1]{Stavros Orfanoudakis}
\credit{Methodology, Writing - review \& editing}

\author[2]{Shengren Hou}
\credit{Writing - review \& editing}

% Third author
\author[1]{Peter Palensky}[%
 ]
%\fnmark[2]
% \ead{cvr3@sayahna.org}
% \ead[URL]{www.sayahna.org}

\credit{Funding acquisition}

% Fourth author
\author%
[1]
{Pedro P. Vergara}[orcid=0000-0003-0852-0169]
\cormark[1]
%\fnmark[1,3]
% \ead{rishi@stmdocs.in}
% \ead[URL]{www.stmdocs.in}

\credit{Conceptualization, Methodology, Writing - review \& editing, Funding acquisition}

\affiliation[2]{organization={Beijing Energy Quant Ltd.},
    city={Haidian, Beijing},
    postcode={53100}, 
    country={China}}

% Corresponding author text
\cortext[cor1]{Corresponding author email: p.p.vergarabarrios@tudelft.nl}

% Footnote text
%\fntext[fn1]{This is the first author footnote. but is common to third
%  author as well.}
%\fntext[fn2]{Another author footnote, this is a very long footnote and
% it should be a really long footnote. But this footnote is not yet
%sufficiently long enough to make two lines of footnote text.}

% For a title note without a number/mark
%\nonumnote{This note has no numbers. In this work we demonstrate $a_b$
%  the formation Y\_1 of a new type of polariton on the interface
%  between a cuprous oxide slab and a polystyrene micro-sphere placed
%  on the slab.
%  }

% Here goes the abstract
\begin{abstract}
Optimal dispatch of energy storage systems (ESSs) in distribution networks involves jointly improving operating economy and voltage security under time-varying conditions and possible topology changes. 
To support fast online decision making, we develop a topology-aware Reinforcement Learning architecture based on Twin Delayed Deep Deterministic Policy Gradient (TD3), which integrates graph neural networks (GNNs) as graph feature encoders for ESS dispatch.
We conduct a systematic investigation of three GNN variants: graph convolutional networks (GCNs), topology adaptive graph convolutional networks (TAGConv), and graph attention networks (GATs) on the 34-bus and 69-bus systems, and evaluate robustness under multiple topology reconfiguration cases as well as cross-system transfer between networks with different system sizes. 
Results show that GNN-based controllers consistently reduce the number and magnitude of voltage violations, with clearer benefits on the 69-bus system and under reconfiguration; on the 69-bus system, TD3-GCN and TD3-TAGConv also achieve lower saved cost relative to the NLP benchmark than the NN baseline.
We also highlight that transfer gains are case-dependent, and zero-shot transfer between fundamentally different systems results in notable performance degradation and increased voltage magnitude violations.
This work is available at: \url{https://github.com/ShuyiGao/GNNs_RL_ESSs} and \url{https://github.com/distributionnetworksTUDelft/GNNs_RL_ESSs}.
\end{abstract}

% Use if graphical abstract is present
% \begin{graphicalabstract}
% \includegraphics{figs/grabs.pdf}
% \end{graphicalabstract}

% Research highlights
\begin{highlights}
\item Topology-aware graph neural network (GNN) encoder embedded in asymmetric actor–critic Reinforcement Learning (RL) representations to capture local actions and system-wide voltage effects for energy storage system (ESS) dispatch.
\item Topology-aware RL reduces voltage-violation frequency and magnitude, with clearer gains on the 69-bus system and under reconfiguration.
\item Transfer benefits are case-dependent: reconfiguration shows no consistent advantage over neural network (NN) RL baselines, and cross-system zero-shot transfer degrades performance.
\end{highlights}

% Keywords
% Each keyword is separated by \sep
\begin{keywords}
Distribution network \sep Energy storage dispatch \sep Voltage regulation \sep Topology reconfiguration \sep Reinforcement learning \sep Graph neural network \sep
\end{keywords}

\maketitle

\begin{figure*}
\begin{mdframed}[linewidth=1pt, skipabove=0pt, skipbelow=0pt, innertopmargin=10pt, innerbottommargin=10pt]
\begin{multicols}{2} % Start two-column environment
\noindent

\begin{tabularx}{\linewidth}{@{}lX@{}}
\multicolumn{2}{@{}l}{\raggedright \underline{\textbf{\emph{Sets}}}} \\ [2pt]
    $\mathcal{N}$ & set of nodes in distribution network \\
    $\mathcal{L}$ & set of lines in distribution network \\
    $\mathcal{B}$ & set of nodes equipped with ESS\\
    $\mathcal{T}$ & set of time steps of MDP \\
    $\mathcal{S}$ & set of states of MDP \\ 
    $\mathcal{A}$ & set of actions of MDP \\
    $\mathcal{P}$ & set of transition probabilities of MDP \\
    $\mathcal{R}$ & set of rewards of RL \\[6pt]
\multicolumn{2}{@{}l}{\raggedright \underline{\textbf{\emph{Indexes}}}} \\[2pt]
    $i$,$j$ & $i$,$j$ th node, $i,j \in \mathcal{N}$\\
    % $ij$ & the $j$ th PV generator, $j \in \mathcal{L}$\\
    $t$ & time step $t \in \mathcal{T}$\\[6pt]
\multicolumn{2}{@{}l}{\raggedright \underline{\textbf{\emph{Parameters}}}} \\[2pt]
    $R_{ij}$ & resistance of line $ij$, $ij \in \mathcal{L}$\\
    $X_{ij}$ & reactance of line $ij$, $ij \in \mathcal{L}$\\
    $\eta_i$ & $i$ th ESS's energy efficiency\\
    $\overline{E_i^B}$ & $i$ th ESS's maximum power\\
    $\overline{SOC}_i^B$, $\underline{SOC}_i^B$ & $i$ th ESS's max/min SOC\\
    $\overline{P}_i^B$, $\underline{P}_i^B$ & $i$ th ESS's max power limit
\end{tabularx}

\begin{tabularx}{\linewidth}{@{}lX@{}}\\
    $\overline{V}$, $\underline{V}$ & max/min voltage magnitude limit\\
    $V_0$ & nominal voltage\\
    $\overline{I}$ &  maximum allowable current magnitude on line $(ij)$\\
    $\varphi_0$, $\varphi_1$ & weight factor of MDP's reward functon\\
    ${\gamma}$ & discount factor of MDP\\
%    ${\alpha}$ & learning rate of RL algorithms\\
%    ${\tau}$ & update rate of DQN's target network\\[6pt]
\multicolumn{2}{@{}l}{\raggedright \underline{\textbf{\emph{Variables}}}} \\[2pt] 
    $P_{i,t}^B$ & ESS's power of $i$ th node at time step $t$\\
    $P_{i,t}^D$ & load demand of $i$ th node at time step $t$\\
    $P_{i,t}^{PV}$ & solar generation of $i$ th node at time step $t$\\
    $P_{ij,t}$ & active power from node $i$ to $j$ at time step $t$\\
    $Q_{ij,t}$ & reactive power from node $i$ to $j$ at time step $t$\\
    $I_{ij,t}$ & current magnitude from node $i$ to $j$ at time step $t$\\
    $V_{i,t}$ & voltage magnitude of node $i$ at time step $t$\\
    $P^S_{i,t}$ & active power injected from the substation into node $i$ at time step $t$\\
    $Q^S_{i,t}$ & reactive power injected from the substation into node $i$ at time step $t$\\
    $SOC_{i,t}^B$ & $i$ th ESS's SOC at time step $t$\\
    
\end{tabularx}
\end{multicols}
\end{mdframed}
\end{figure*} 

\section{Introduction}

The increasing penetration of distributed energy resources (DERs), together with time-varying electricity prices, has made real-time operation of distribution networks (DNs) increasingly challenging \cite{azarnia2024offering}. 
High penetration of DERs introduces uncertainty, variability, and bidirectional power flows, while price volatility requires timely decisions because economically preferred actions can change over short intervals \cite{xing2023real}.
In particular, distribution network operation must jointly maintain energy balance, reduce operating cost, and satisfy voltage and device constraints, which makes practical operation a tightly coupled constrained dispatch problem that co-optimizes economic objectives with network security \cite{shengren2023optimal}.
Energy storage systems (ESSs) provide a key source of flexibility in distribution networks by enabling temporal energy shifting, smoothing net-load fluctuations, and supporting local voltage regulation \cite{jaradat2025review}. 
Accordingly, ESS dispatch is commonly formulated as an optimization problem that coordinates charging/discharging decisions with network and device constraints, such as voltage limits, line constraints, and state-of-charge (SOC) dynamics, which often leads to non-linear programming (NLP) formulations. 
Model-based optimizations, such as stochastic programming \cite{vergara2020stochastic,Rev_MINLP2,garcia2021stochastic} and robust optimization \cite{lu2021multistage,xiong2024two} can deliver high-quality solutions; however, repeatedly solving them in real time can be computationally demanding, particularly for large networks or when operating conditions change frequently \cite{li2024optimal}.

The growing availability of data from smart meters and sensors further enhances the viability of data-driven Reinforcement Learning (RL) as an alternative for operational decision making \cite{gao2024linear}.
The optimization problem of ESS dispatch can be formulated as a Markov decision process (MDP), and solved using powerful data-driven RL algorithms. In detail, by training offline with a simulator and executing online with a forward pass of the policy network, RL can provide near-instant control actions, alleviating the latency bottleneck of repeatedly solving complex optimizations \cite{Rev_3}. 
Accordingly, RL has been increasingly explored for distribution network operation, including ESS dispatch \cite{gao2025symbolic,deng2025dr,xu2024optimal} and voltage regulation \cite{hou2024distflow,xiang2023deep}, and open-source toolkits \cite{henry2021gym,hou2024rl} have been developed to facilitate research and benchmarking.

Despite these advances, two challenges remain in distribution networks.
First, voltage regulation is topology-dependent: the impact of ESS actions on voltages is governed by electrical connectivity and line impedances, and becomes harder to capture in deeper, more heterogeneous networks \cite{licari2025addressing}. 
Second, distribution networks frequently experience topology reconfiguration due to switching operations and service restoration, which alters coupling among buses \cite{abu2018modern}. 
As a result, topology-agnostic controllers may generalize poorly and yield inconsistent constraint satisfaction, motivating graph-structured learning models.

Distribution networks naturally form graphs of buses and lines, making graph neural networks (GNNs) a natural choice for representation learning \cite{solheim2024visualizing}. 
By performing message passing over the grid topology, GNNs learn graph embeddings from node features, which motivates the following advantages over topology-agnostic neural networks(NNs):
\begin{itemize}
\item \textbf{Spatial awareness of physical relationships}\\
    GNNs inherently capture spatial dependencies and local interactions between physically connected nodes, which aligns with the network nature and physical constraints (e.g., power flow) of energy systems \cite{hossain2025topology}.
    \item \textbf{Efficient learning on large networks}\\
    Compared with topology-agnostic NNs that rely on flattened states, GNNs leverage sparse connectivity and shared local aggregation, which becomes increasingly advantageous in larger and more heterogeneous networks by preserving topology information without a dense representation \cite{lin2024powerflownet}.
    \item \textbf{Transferability across heterogeneous networks}\\
    GNNs have the potential to generalize to unseen or reconfigured topologies under consistent node feature definitions \cite{donon2024topology}, whereas NN-based controllers built on fixed-size representations are typically tied to a specific input dimension and can be less flexible when the network size changes.
\end{itemize}

This makes GNNs especially suitable for RL-based ESSs dispatch, and has motivated the recent development of topology-aware RL frameworks.
For example, \cite{xing2023real} proposed a graph-based RL framework for real-time optimal scheduling in active distribution networks, using a graph attention network (GAT) module to encode topology-aware states and classic actor-critic based RL deep deterministic policy gradient (DDPG) to generate control actions. 
Experiments on a modified IEEE 33-bus system under fault scenarios showed improved scheduling performance and adaptability to topology variations.
Similarly, \cite{chen2023physical} proposes an RL approach with a GAT-based topology encoder for voltage regulation in distribution networks.
Tests on IEEE 33-node and 136-node systems demonstrated faster convergence and better voltage control performance than existing RL baselines, while a GNN-RL based dynamic reconfiguration optimization model for energy storage management under faults and varying network scenarios has been proposed in \cite{gao2025graph}. 
Case studies reported improved decision accuracy, reduced voltage violations, and enhanced operational security.
In \cite{wu2023two}, a two-stage voltage regulation framework is presented, combining day-ahead mixed-integer second-order cone optimization programming (MISOCP) and a real-time graph convolutional network (GCN) based RL controller for fast voltage magnitude support. 
Simulations showed effective mitigation of voltage magnitude fluctuations with real-time applicability.
Nevertheless, the above-mentioned works typically adopt a single GNN architecture and rarely provide systematic comparisons of different GNN variants against standard NN baselines or systematic transfer tests across multiple topology reconfiguration cases.
For practical deployment, an RL controller should jointly deliver strong economic performance, minimize voltage magnitude violations, and be robust to topology changes.
This motivates a comprehensive evaluation of GNN-based RL for ESS optimal dispatch, through consistent comparisons across GNN architectures, network sizes, and reconfiguration scenarios.

To this end, this paper introduces a topology-aware GNN-based RL architecture for optimal dispatch of ESSs in distribution networks, thoroughly evaluated under both economic and voltage-related metrics.
The main contributions are summarized as follows:
% \begin{itemize}
%     \item We developed GNN-based actor-critic policies for distribution network ESS optimal dispatch, where graph feature encoders explicitly incorporate network topology and electrical coupling.
%     \item We conducted a comprehensive comparison across multiple GNN architectures (GCN, TAGConv, and GATConv) and a standard NN baseline under consistent training settings, highlighting performance and stability trade-offs.
%     \item We evaluateed generalization across two benchmark systems (34-bus and 69-bus) and systematically test transfer under multiple topology reconfiguration cases, as well as cross-system transfer between networks with different size.
% \end{itemize}
\begin{itemize}
    \item We develop a topology-aware graph feature encoder for ESS dispatch, and integrate three representative GNNs: graph convolutional networks (GCN), topology-adaptive graph convolutional networks (TAGConv), and graph attention networks (GAT), to extract embeddings that incorporate network topology and electrical coupling.
    \item We integrate the graph feature encoder into TD3 with asymmetric actor-critic representations: the actor uses ESS-focused embeddings derived from the battery nodes for action generation, while the critic aggregates system-level information via global pooling to estimate rewards under network constraints.
    \item We systematically assess robustness and generalization under topology changes, including multiple topology reconfiguration cases within each system and cross-system transfer between networks of different system sizes.
\end{itemize}

The remainder of this paper is organized as follows. 
Section~\ref{sec:math} models the ESS optimal dispatch problem as an NLP formulation and formulates it in to an MDP.
Section~\ref{sec:gnn} introduces the proposed GNN-based RL framework, including the proposed graph feature encoder and the actor-critic training procedure.
Section~\ref{sec:results} presents experimental settings and evaluation metrics, discusses the results on the 34-bus and 69-bus systems, including convergence, operational behavior, voltage magnitude regulation, and topology reconfiguration transfer.
Finally, Section~\ref{sec:conclusion} concludes the paper and outlines future work.

\section{Mathematical Formulation and MDP for Optimal ESSs Dispatch}\label{sec:math}
\subsection{Mathematical formulation}
In a distribution network with buses equipped with ESSs, the optimal ESS dispatch is formulated as an NLP problem with the objective to minimize total operational costs while satisfying grid-level constraints.
The optimization objective, defined in (\ref{eq:op_obj}), minimizes the cost of the electricity purchase from the main grid over the scheduling horizon ${\cal T}$, where the time-varying price $\sigma_t$ reflects real-time supply-demand imbalances. This optimization is subject to the constraints given in (\ref{eq:active_power_balance}) – (\ref{eq:power_not_substation}).

\begin{equation}\label{eq:op_obj}
    \min_{\substack{P^{B}_{i,t}, \forall i \in {\cal B}, \forall t \in {\cal T}}} \left\lbrace  \sum_{t \in {\cal T}} \left[ \sigma_{t}\sum_{i \in {\cal N}} \left(P^D_{i,t} - P^{PV}_{i,t} - P^{B}_{i,t}\right)\Delta t \right] \right\rbrace
\end{equation}
s.t.
\vspace{-2mm}
\begin{multline} \label{eq:active_power_balance}
 \hspace{-5mm} \sum_{ji \in {\cal L}} P_{ji,t} - \sum_{ij \in {\cal L}} (P_{ij,t} + R_{ij}I_{ij,t}^2) +P_{i,t}^{B} \\ + P_{i,t}^{PV}+ P_{i,t}^{S}= P_{i,t}^{D}  \quad \forall i \in {\cal N}, \forall t \in {\cal T}  
\end{multline}
\vspace{-6mm}
\begin{multline} \label{eq:reactive_power_balance}
 \hspace{-5mm} \sum_{ji \in {\cal L}} Q_{ji,t} - \sum_{ij \in {\cal L}} (Q_{ij,t} + X_{ij}I_{ij,t}^2) + Q_{i,t}^{S} = Q_{i,t}^{D}  \\  \forall i \in {\cal N}, \forall t \in {\cal T}
\end{multline}
\vspace{-6mm}
\begin{multline}\label{eq:votlage_drop}
 \hspace{-5mm} V_{i,t}^2-V_{j,t}^2=2(R_{ij}P_{ij,t}+X_{ij}Q_{ij,t}) + (R_{ij}^2+X_{ij}^2)I_{ij,t}^2 \\ \quad  \forall i,j \in {\cal N}, \forall t \in {\cal T}  
\end{multline}
\vspace{-8mm}
\begin{multline}\label{eq:vi=pq}   
 \hspace{-5mm}V_{i,t}^2I_{ij,t}^2=P_{ij,t}^2+Q_{ij,t}^2 \quad \quad \quad \quad  \forall i,j \in {\cal N}, \forall t \in {\cal T} 
\end{multline}
\vspace{-8mm}
\begin{multline}\label{eq:voltage_boundary}   
 \hspace{-5mm}\underline{V}^{2}\leq V_{i,t}^2\leq \overline{V}^{2} \quad \quad \quad \quad \quad \quad \quad \forall i \in {\cal N}, \forall t \in {\cal T}
\end{multline}
\vspace{-8mm}
\begin{multline}\label{eq:limites_corre_5}   
 \hspace{-5mm} 0 \leq I_{ij,t}^2 \leq \overline{I}_{ij}^{2} \quad \quad \quad \quad \quad \quad \quad  \forall ij \in {\cal L}, \forall t \in {\cal T}
\end{multline}
\vspace{-8mm}
\begin{multline}\label{eq:power_not_substation}   
 \hspace{-5mm} P_{i,t}^{S} = Q_{i,t}^{S} = 0 \quad \quad \quad \quad \quad  \forall i \in {\cal N} \backslash \{1\}, \forall t \in {\cal T}
\end{multline}
\vspace{-8mm}
\begin{multline} \label{eq:SOC_cha}
\hspace{-5mm} SOC_{i,t}^{B}=SOC_{i,t-1}^{B} - \eta^{B}_{i}P_{i,t}^{B}\Delta t/\overline{E}^{B}_{i} \quad \forall i \in {\cal{B}}, \forall t \in {\cal{T}}
\end{multline}
\vspace{-6mm}
\begin{flalign}
 \hspace{4mm} & \underline{SOC}_{i}^{B}\leq SOC_{i,t}^{B}\leq\overline{SOC}_{i}^{B} & \forall i \in {\cal{B}}, \forall t \in {\cal{T}} & \label{eq:SOC_cons}\\
& \underline{P}^{B}_{i}\leq P^{B}_{i,t}\leq \overline{P}^{B}_{i} & \forall i \in {\cal B}, \forall t \in {\cal T} & \label{eq:char_disc_cons}
\end{flalign}

To accurately represent physical and operational constraints, the grid-level constraints are modeled via a detailed power flow formulation, comprising equations~\eqref{eq:active_power_balance}–\eqref{eq:vi=pq}. These constraints govern the network's active power flow $P_{ij,t}$, reactive power $Q_{ij,t}$, current magnitude $I_{ij,t}$ on each line $(i,j)$, and voltage magnitude $V_{i,t}$ at each node $i$. Voltage safety is ensured through \eqref{eq:voltage_boundary} and \eqref{eq:limites_corre_5}, which enforce permissible voltage bounds and current limits, respectively. 
Additionally, \eqref{eq:power_not_substation} restricts substation connection to a single designated node, reflecting typical radial distribution system configurations.
The power flow constraints ensure that the ESS scheduling decisions respect grid-level operational feasibility, including compliance with node voltage and line current magnitude limits.

The operational constraints of ESSs are captured in equations~\eqref{eq:SOC_cha}–\eqref{eq:char_disc_cons}. Specifically, \eqref{eq:SOC_cha} models the SOC evolution of each ESS that is dynamically updated based on the charging and discharging power, where a positive value of $P_{i,t}^B$ represents ESS is discharging.
Expression~\eqref{eq:SOC_cons} ensures that the SOC of each ESS remains within predefined operational limits to preserve battery health and ensure availability. 
Finally, constraint~\eqref{eq:char_disc_cons} imposes upper bounds on charging and discharging power, limiting the instantaneous power flow into and out of the storage units. 
$\overline{P}^{B}_{i}$ is a positive value donating the maximum discharging power, and $\underline{P}^{B}_{i}$ is negative as maximum charging power.

\subsection{ESSs optimal dispatch as a Markov Decision Process}
The optimal energy dispatch problem of ESSs can be formulated as an MDP, defined by a 5-tuple ($\mathcal{S}, \mathcal{A}, \mathcal{P}, \mathcal{R}, \gamma$), where $\mathcal{S}$ is the state space, $\mathcal{A}$ is the action space, $\mathcal{P}$ represents the state transition probabilities, $\mathcal{R}$ is the reward function, and $\gamma$ is the discount factor.
In reinforcement learning, the goal is to learn a policy $\pi(a|s)$ that maximizes the expected discounted cumulative reward over a finite horizon~$\mathcal{T}$, i.e.,
$\max_{\pi}\ \mathbb{E}_{\pi}\!\left[\sum_{t=0}^{\mathcal{T}-1}\gamma^{t} r_t\right]$, where $r_t=\mathcal{R}(s_t,a_t,s_{t+1})$.
As defined in~(\ref{state}), the state $s_t$ characterizes the operational condition of the distributed energy system at time $t$, comprising the following variables: 
current timestep $t$, electricity price $\sigma_t$, active power demand at each node $P_{i,t}^D$, state of charge of ESS-equipped nodes $SOC_{i,t}^B$, and node voltage $V_{i,t} (i\in \mathcal{B})$. 
Noted that here we are only considering SOC and voltage information from the battery nodes instead of all nodes. 
These variables reflect distinct attributes of the system. 
Specifically, $t$, $P_{i,t}^D$, and $V_{i,t}$ are deterministic and follow predictable physical laws or scheduling patterns; $\sigma_t$ and $P_{i,t}^D$ include stochastic components arising from market volatility and user behavior; $SOC_{i,t}^B$ represents a controllable variable directly influenced by dispatch decisions. 
The action $a_t$, as specified in~(\ref{action}), denotes the charging or discharging power of each ESS $i\in \mathcal{B}$, serving as the key decision variable that directly affects system performance by shaping energy flows, balancing supply and demand, and optimizing operational objectives.
\begin{equation}
\label{state}
    s_t = \{t, \sigma_{t},P_{i,t}^D|_{i \in \mathcal{N}}, SOC_{i,t}^B |_{i \in \mathcal{B}},V_{i,t}|_{i \in \mathcal{B}}\}, s_t\in \mathcal{S}
\end{equation}    
\vspace{-4mm}
\begin{equation}
\label{action}
a_t = \{P_{i,t}^B|_{i \in \mathcal{B}}\}, a_t \in \mathcal{A}
\end{equation}

The transition probability is denoted by $\mathcal{P}: {s_t, a_t} \in \mathcal{S} \times \mathcal{A} \rightarrow p \ \dot{=} \ p(s_{t+1} | s_t, a_t)$ and characterizes the system’s internal dynamics by defining the likelihood of reaching state $s_{t+1}$ from state $s_t$ after taking action $a_t$.
$\mathcal{R}$ is the set of reward function where $\mathcal{R}: {s_t, a_t, s_{t+1}} \in \mathcal{S} \times \mathcal{A} \times \mathcal{S} \rightarrow r \ \dot{=} \ \mathcal{R}(s_t, a_t, s_{t+1}) \in \mathbb{R}$, which evaluates the quality of an action under a predefined objective, providing feedback to guide the learning process.
The discount factor $\gamma \in [0,1]$ controls the agent’s preference for immediate versus long-term rewards. A value of $\gamma=1$ allows the agent to fully account for future consequences, while $\gamma=0$ leads it to focus solely on immediate gains.
\begin{equation}
\label{reward}
r_t=\varphi_0 r_0 + \varphi_1r_1
\end{equation}

In the formulated MDP, the reward function is designed to guide the RL agent toward an optimal policy that jointly minimizes electricity costs from the external grid and maintains voltage stability within the distribution network.
As defined in~(\ref{reward}), the reward consists of two components weighted by coefficients $\varphi_0$ and $\varphi_1$, which allow system operators to tune the agent’s priorities between economic and technical objectives.
\begin{equation}
\label{rforpower}
r_0 = \sigma_t[\sum_{i\in\mathcal{B}}P_{i,t}^B]\Delta t
\end{equation}

The first component, $r_0$ in~(\ref{rforpower}), measures the monetary benefit obtained by discharging ESSs rather than purchasing electricity from the grid at the prevailing price $\sigma_t$. A higher $r_0$ reflects more profitable dispatch behavior and better alignment with price signals.
\begin{equation}
r_1=\sum_{i\in\mathcal{B}}\{min\{0,(\frac{\overline{V}-\underline{V}}{2}-|V_0-V_{i,t}|)\}\}
\label{rforpen}
\end{equation}

The second component, $r_1$ in~(\ref{rforpen}), imposes a penalty for voltage deviations. $r_1$ quantifies how far the voltage $V_{i,t}$ at each ESS node deviates from a nominal value $V_0$, with penalties applied when voltages fall outside the allowable range $[\underline{V}, \overline{V}]$. This term encourages the agent to maintain voltage levels within operational limits, promoting system stability.

\section{GNN-Based Reinforcement Learning for Optimal ESSs Dispatch}\label{sec:gnn}

\begin{figure*}[htbp]
    \centering
    \includegraphics[width=2.0\columnwidth]{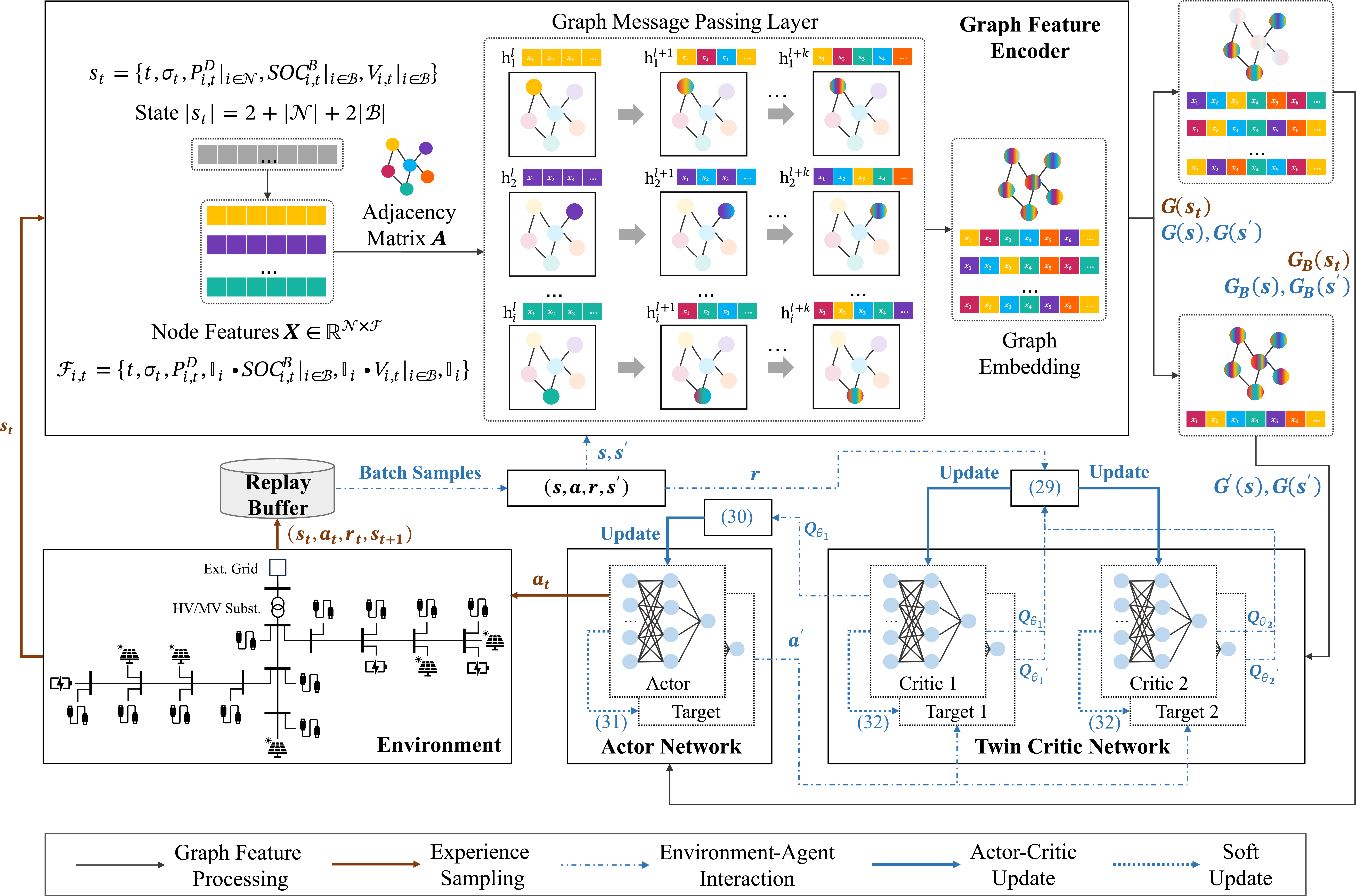}
    \caption{Architecture of the proposed GNN-based RL framework for distribution-network ESS optimal dispatch, consisting of the graph feature encoder, the ESS dispatch environment, the experience replay buffer, and the actor–critic networks.}
    \label{fig:Struc}
\end{figure*}

An overview of the proposed GNN-based RL framework is illustrated in Fig.~\ref{fig:Struc}.
We design a GNN encoder that learns a representation combining distribution network bus attributes and graph topology. 
This representation informs the actor-critic networks, which parameterize the policy and value functions. 
By interacting with the environment, the agent refines this policy to exploit graph-structured information for sequential decisions, ultimately learning to maximize long-term reward.

\subsection{Graph Feature Encoder for Distribution Network State}
Distribution networks are fundamentally structured like graphs, with buses and lines forming an interconnected system.  
GNNs use a repeated message-passing process, where each node updates its own representation by gathering information from its electrically connected neighbors. This captures the local physical relationships and spatial dependencies that are critical for maintaining stable power flow.
Importantly, GNNs learn from node features together with the graph structure, rather than from a fixed-size flattened input.
This design can facilitate generalization to grids with different sizes, layouts, or structural changes. 
These advantages make GNNs a powerful and natural choice for modeling and learning distribution networks.

To effectively encode the graph-structured features of a state $s_t$ and leverage the capability of GNNs to aggregate information from neighboring nodes, we design a graph feature encoder, as illustrated in Fig.~\ref{fig:Struc}.
Each state is represented as a graph $\mathbf{G} = (\mathbf{E}, \mathbf{X})$, where $\mathbf{E} \in \mathbb{R}^{\mathcal{N} \times \mathcal{N}}$ is the adjacency matrix of the distribution network, indicating the connectivity between nodes. Specifically, $\mathbf{E}_{ij} = 1$ if there exists a connection between node $i$ and node $j$, and $0$ otherwise.
$\mathbf{X} \in \mathbb{R}^{\mathcal{N} \times \mathcal{F}}$ denotes the node feature matrix, where each row corresponds to the features of a node that are derived from the original state vector $s_t$.
In a conventional neural network setting, $s_t$ is represented as a long flat vector, whose dimension scales with the number of nodes and ESS-connected nodes (i.e., $|s_t| = 2+ |\mathcal{N}| + 2|\mathcal{B}| $ in our formulation).

In contrast, the GNN-based encoder represents $s_t$ as a set of node-level feature vectors.
For each node $i\in\mathcal{N}$, we construct a feature vector $\mathcal{F}_{i,t}$ from its intrinsic state information and a binary indicator $\mathbb{I}_i$ in (\ref{bi}) that specifies whether node $i$ is connected to an ESS.
Here, same as in (\ref{action}), $V_{i,t}$ denotes the voltage magnitude measurement provided by the environment only at ESS-connected nodes.
We adopt zero-padding \cite{rossi2022unreasonable} as a simple baseline to maintain a fixed-dimensional node feature representation when ESS-specific variables are unavailable.
For non-ESS nodes, $\{SOC_{i,t}^B, V_{i,t}\}$ are not defined in the state and thus unobserved; their corresponding feature entries are padded with zeros.
The binary indicator $\mathbb{I}_i$ is included to explicitly encode the availability of these ESS-specific variables, preventing the model from interpreting padded zeros as meaningful physical values.
Accordingly, the node feature vector is defined as in (\ref{rawFea}).
\begin{equation}\label{bi}
    \mathbb{I}_i =
\begin{cases}
1, & i \in \mathcal{B},\\
0, & \text{otherwise}.
\end{cases}
\end{equation}
\begin{equation}\label{rawFea}
\mathcal{F}_{i,t} =
\Big\{t,\ \sigma_t,\ P_{i,t}^D,\ \mathbb{I}_i\cdot\,SOC_{i,t}^B,\mathbb{I}_i\cdot\,V_{i,t},\ \mathbb{I}_i\Big\},
 \forall i\in\mathcal{N}
\end{equation}

Once the graph $\mathbf{G} = (\mathbf{E}, \mathbf{X})$ is constructed, it is passed through a sequence of message passing layers to iteratively refine the node representations by aggregating information from their neighbors. 
In each layer $l\in \mathbf{L}$, the update can be written in the general form:
\begin{equation}
    \mathbf{H}^{l+1}=f(\mathbf{H}^{l},\mathbf{E}), l\in\mathbf{L}
\end{equation}
where $\mathbf{H}^{l}$ denotes the node feature matrix at layer $l$, $\mathbf{H}^{0}$ is identical to $\mathbf{X}$, and $f(\cdot)$ is the message aggregation function defined by the specific GNN variant.
Intuitively, during message passing, each node collects features from its directly connected neighbors and combines them with its own features, enabling the network to capture both local topology and feature interactions. 
As shown in Fig.~\ref{fig:Struc}, different colors represent the features of different nodes that are gradually blended through successive layers, illustrating how the information diffuses across the graph.
After several rounds of message passing, node features are transformed into graph embeddings $\mathbf{G}(s_t)$, which further serves as inputs to the actor and critic networks.

The actor network extracts and aggregates graph embeddings from ESS-related nodes to form a compact battery-level representation $\mathbf{G}_B(s_t)$. 
This design enables the policy to focus on storage-specific features, ensuring a lightweight and computationally efficient architecture. 

For critic networks, the graph embeddings are transformed into a graph-level representation $\mathbf{G}^{'}(s)$ using global mean pooling. This operation takes the average of all node embeddings in the graph:
\begin{equation}
    \mathbf{h}_{graph}=\frac{1}{\mathcal{N}}\sum_{i=1}^\mathcal{N}\mathbf{h}_i
\end{equation}
where $\mathbf{h}_i$ is the feature vector of node $i$. 
Global mean pooling plays a crucial role in summarizing the overall state of the system while remaining invariant to node ordering, ensuring that the learned graph representation captures both the aggregated structural information and the distribution of node features.

Moreover, given that $\mathbf{G}_B(s_t)$ and $\mathbf{G}^{'}(s)$ are derived from the learned graph structure rather than flattening all node features into a single long vector as in \cite{9906438}, it can be naturally transferred across diverse network topologies and operating scenarios with minimal adjustments.
This property is particularly valuable for distribution systems, where the network configuration may change due to reconfiguration or expansion. 

\subsection{Graph Massage Passing Layer}\label{message_passing_layers}

To implement the message-passing function $f(\cdot)$, we we employ three representative graph convolutional architectures: GCNConv\cite{kipf2016semi}, TAGConv\cite{du2017topology}, and GATConv\cite{velivckovic2017graph}.
These layers differ in how they aggregate neighbor information, their sensitivity to graph topology, and their ability to capture long-range dependencies.
The following subsections introduce the principle, layer update formulation, and advantages of each approach.

\subsubsection{Graph Convolutional Network}
The GCNConv layer aggregates information from neighboring nodes using a normalized adjacency matrix to ensure stable propagation:
\begin{equation}
    \mathbf{h}_i^{l+1} = \phi\!\left( 
\sum_{j \in \mathcal{M}(i) \cup \{ i \}} 
\frac{1}{\sqrt{\hat{d}_i \hat{d}_j}} \; 
\mathbf{W}^{l} \mathbf{h}_j^{l} 
\right)
\end{equation}
where $\mathbf{h}_i^{l}$ is the feature vector of node $i$ at layer $l$, $\mathbf{W}^{l}$ is the learnable weight matrix, $\hat{d}_i$ is the degree of node $i$ after adding self-loops, $\mathcal{M}(i)$ denotes the neighbors of $i$, and $\phi(\cdot)$ is a nonlinear activation function (e.g., ReLU).

By applying symmetric normalization, GCNConv achieves computational efficiency and avoids numerical instability. This formulation is particularly suited for graphs with fixed topology, as in power systems, where stable neighbor aggregation can exploit the structural regularity of the network.

\subsubsection{Topology Adaptive Graph Convolution}
TAGConv extends GCNConv by incorporating K-order polynomial filters of the adjacency matrix, enabling multi-hop aggregation in a single layer:

\begin{equation}
    \mathbf{h}_i^{l+1} = \phi\!\left( 
\sum_{k=0}^{K} \sum_{j \in \mathcal{M}_k(i)} 
\mathbf{W}_k^{l} \mathbf{h}_j^{l} 
\right)
\end{equation}
where $\mathcal{M}_k(i)$ is the set of nodes at exactly $k$ hops from node $i$, $\mathbf{W}_k^{l}$ is the learnable weight matrix for the $k$-hop filter, and $K$ is the maximum hop order.

By allowing different weights for different hop distances, TAGConv captures both local and global structure without stacking many layers. 
This is advantageous in systems where the effect of control actions may propagate across multiple hops, reducing the need for deep architectures and mitigating over-smoothing.
%cite nan&stavros's paper

\subsubsection{Graph Attention Network}
GATConv introduces an attention mechanism that assigns different weights to different neighbors, enabling the model to focus on the most relevant connections. A variant of classical GATConv\cite{brody2021attentive} is adopted, and the attention score for edge $(i,j)$ is computed as:
% \begin{equation}
%     e_{ij} = \text{LeakyReLU}\!\left( 
% \mathbf{a}^\top \!\left[ 
% \mathbf{W}^{l} \mathbf{h}_i^{l} \,\|\, 
% \mathbf{W}^{l} \mathbf{h}_j^{l} 
% \right] 
% \right)
% \end{equation}
\begin{equation}
    e_{ij} = \mathbf{a}^\top \!\text{LeakyReLU}\!\left( 
\mathbf{W}^{l} \left[\mathbf{h}_i^{l} \,\|\, 
 \mathbf{h}_j^{l} 
\right] 
\right)
\end{equation}
where $\mathbf{a}$ is a learnable attention vector, $\|\,$ denotes vector concatenation, and $\mathbf{W}^{l}$ is the shared weight matrix.
The normalized attention coefficient is:
\begin{equation}
    \alpha_{ij} = \frac{\exp(e_{ij})}{\sum_{k \in \mathcal{M}(i)} \exp(e_{ik})}
\end{equation}
and the node update is given by:
\begin{equation}
    \mathbf{h}_i^{l+1} = \phi\!\left( 
\sum_{j \in \mathcal{M}(i)} 
\alpha_{ij} \mathbf{W}^{l} \mathbf{h}_j^{l} 
\right)
\end{equation}
where $\alpha_{ij}$ is the normalized attention coefficient and $\phi(\cdot)$ is a nonlinear activation function (e.g., ReLU).

This formulation allows the model to adaptively emphasize influential neighbors and suppress less relevant ones. It is particularly effective in heterogeneous graphs, such as electrical networks with diverse node roles, such as storage units, loads, generators, and it offers robustness to topology variations.

\subsection{Reinforcement Learning Algorithm}
% \textcolor{red}{[here in this section, I will skip some most basic RL knowledge, will introduce in the introduction instead]}

To handle continuous actions with improved stability and sample efficiency, we employed TD3 \cite{fujimoto2018addressing}, which uses a replay buffer to reuse past transitions, thereby improving sample efficiency during training. This is beneficial in graph-structured environments where generating new rollouts can be costly. 
The TD3 training workflow is illustrated in Fig.~\ref{fig:Struc}. 

The brown flows illustrate the transition sampling.
At each time step, the environment returns a state $s_t$, which is encoded into a graph embedding $\mathbf{G}(s_t)$ via the graph feature encoder. 
An action is given consisting of a deterministic action that is obtained from the actor network $\pi_{\mu}$ and an exploration noise $\epsilon_t$, as in (\ref{actor_action}). 
Actor network takes graph features from $\mathbf{G}_B(s_t)$, which provides a compact battery-centric representation for decision making.
The exploration noise is added to encourage sufficient exploration in the continuous action space and to prevent premature convergence to suboptimal deterministic policies.
\begin{equation}\label{actor_action}
    a_t=\text{clip}(\pi_{\mu}(\mathbf{G}_B(s_t))+\epsilon_t,-1,1),  \epsilon_t \sim \mathbb{N}(0,\xi^2)
\end{equation}
The action $a_t$ is executed in the environment to obtain next state $s_{t+1}$ and reward $r_t$.
The transition tuples of $(s_t,a_t,r_t,s_{t+1})$ are then stored in the replay buffer $\mathcal{D}$.

During training, as in blue flows, a mini-batch of transitions $(s,a,r,s')$ are sampled from replay buffer, through graph feature encoder, the pooled graph embeddings $\mathbf{G}'(s)$ and $\mathbf{G}'(s')$ are obtained for the update of twin critic network $Q_{\theta_1}$ and $Q_{\theta_2}$.
To compute a stable target, the target actor network $\pi_{\mu'}$ is used and smoothed:
\begin{equation}\label{actor_action_update}
    a'=\text{clip}(\pi_{\mu'}(\mathbf{G}_B(s'))+\varepsilon,-1,1),  \varepsilon =\text{clip} (\mathbb{N}(0,{\tilde{\xi}}^2),-c,c)
\end{equation}
Here, a clipped Gaussian noise $\varepsilon$ is added to the target action to perform target policy smoothing, which makes the target value less sensitive to small perturbations of the action and improves robustness against function approximation errors.
The target Q-values are then defined using the minimum of the two target critics $Q_{\theta_1'}$ and $Q_{\theta_2'}$, which helps alleviate overestimation bias:
\begin{equation}
    y=r+\gamma \min_{m=1,2} Q_{\theta_i'}(\mathbf{G}'(s'),a')
\end{equation}
Both critics are optimized by minimizing the mean-squared Bellman error:
\begin{equation}
    L(\theta_m)=\mathbb{E}_{(s,a,r,s')\sim \mathcal{D}} [Q_{\theta_m}-y]^2, m\in \{1,2\}
\end{equation}
The actor network is updated less frequently than the critics. 
Specifically, every $n$ critic updates, the actor is optimized to maximize the critic’s estimated value by gradient ascent:
\begin{equation}
    J(\mu)=\mathbb{E}_{s\sim \mathcal{D}}[Q_{\theta_1}(\mathbf{G}_B(s),\pi_{\mu}(\mathbf{G}_B(s)))]
\end{equation}
Finally, the target networks of the actor and critics are updated using soft updates \cite{lillicrap2015continuous}, allowing them to track the online networks smoothly and improving training stability as in (\ref{softupdate_actor}) and (\ref{softupdate_actor}), where $\tau\in(0,1)$ controls the update rate.
\begin{equation}\label{softupdate_actor}
    \mu'\leftarrow\tau\mu+(1-\tau)\mu'
\end{equation}
\begin{equation}\label{softupdate_critic}
    \theta_m'\leftarrow\tau\theta_m+(1-\tau)\theta_m', m\in{1,2}
\end{equation}

\section{Case Study}\label{sec:results}
The experiments are conducted in an open-source environment \cite{hou2024rl} that simulates the dispatch of ESSs in distribution networks. 
A day-ahead operation horizon with a 15-min resolution is considered, resulting in 96 decision steps for a single operating day. 
To investigate the effectiveness of GNNs in ESS energy dispatch and voltage magnitude regulation, two benchmark distribution systems, i.e., the 34-bus system and the 69-bus system, are adopted. 
These two networks differ in topology complexity and scale, enabling a systematic evaluation of the learning performance and generalization capability of the proposed GNN-based policy under different grid conditions. 
The ESSs are set on nodes $\{12,16,27,30,34\}$ in the 34-bus system, and are modeled with an energy capacity of $500$~kWh and a maxium charging/discharging power of $200$~kW. 
For the 69-bus system, ESSs are set on nodes $\{14,16,18,20,22,24,26,27,65\}$, the energy capacity is $1000$~kWh and a power rating of $300$~kW. 
In both cases, the charging/discharging efficiencies are set to $\eta_i=1$, and the SOC is constrained within $[0.2,0.8]$ with initial SOC as 0.4.

All training experiments were performed on the Delft Blue high-performance computing (HPC) system \cite{DHPC2024}. 
To reduce stochastic bias and improve reproducibility, each setting is repeated with five random seeds, and the reported results are averaged across seeds. 
All the RL algorithms share the same setting of hyperparameters with a learning rate of $6e-5$, discount factor $\gamma=0.995$, a replay buffer size of 1e6, a batch size of $512$, and soft target updates with $\tau=5e-3$. 
Target policy smoothing is enabled with noise scale $0.1$, and the actor is updated every $2$ critic updates.

For the TD3-NN, the actor and critic adopt the same multilayer perceptrons (MLPs) size with 3 fully-connected layers and 256 hidden units per layer (ReLU activations), and the critic follows the standard twin-Q design.
For the GNN-based TD3, the actor and critic are implemented as two separate graph networks with the same depth. 
Both networks consist of 3 graph layers, where the hidden dimensions are set to $64$. 
For the actor, the ESS nodes' embeddings are mapped to actions by an MLP with 2 hidden layers of 256-units, followed by a $\mathbb{\tanh}(\cdot)$ output to bound actions in $[-1,1]$. 
For the critic, actions are concatenated to node features as an extra channel, and the pooled graph representation is fed into a 2 hidden layers of 256-units MLP with twin linear heads to estimate $Q_{\theta_1}(s,a)$ and $Q_{\theta_2}(s,a)$.

For performance benchmarking, RL policies are compared against a global optimization baseline implemented in Pyomo, where the ESS dispatch problem is formulated as an NLP problem and solved using \textit{IPOPT} \cite{wachter2006implementation}. 
The NLP is solved offline over the full scheduling horizon and assumes perfect knowledge of the future input profiles (e.g., day-ahead load/net-load and price trajectories), which provides an optimistic reference under the assumed model accuracy.
This baseline serves as a reference upper bound under the assumed model accuracy, allowing a quantitative comparison in terms of saved costs and voltage magnitude constraint satisfaction.

\subsection{Performance of RLs on Training Topology}

\begin{figure}[htbp]
    \centering
    \includegraphics[width=1.0\columnwidth]{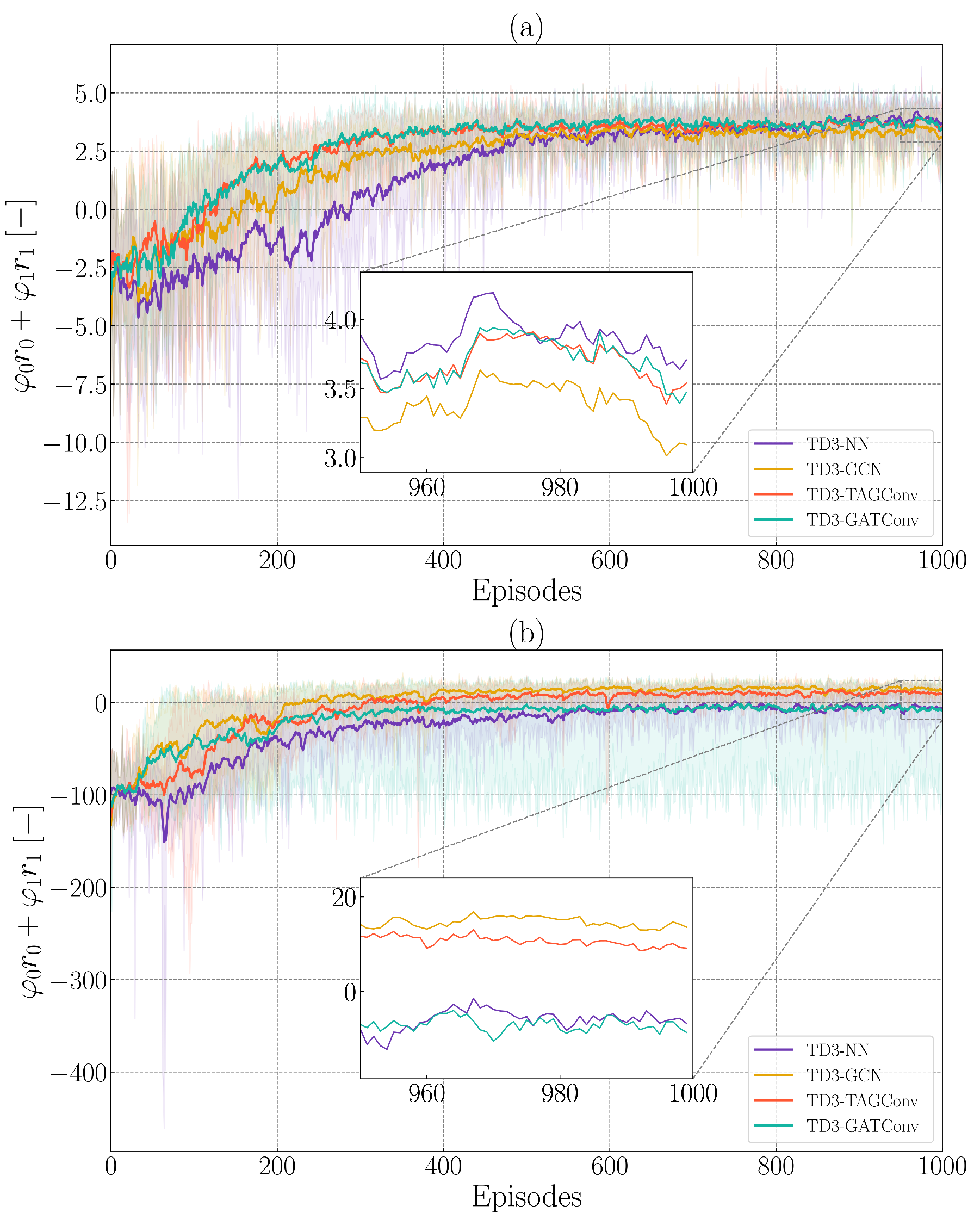}
    \caption{The average reward of RL algorithms during 1000 episodes of training on (a)34 buses system, (b)69 buses system.}
    \label{fig:conver}
\end{figure}
Fig.~\ref{fig:conver} presents the training trajectories of all RL algorithms on the 34-bus and 69-bus systems over 1000 episodes. 
On the 34-bus system, the episode reward increases from approximately $-2.5$ to $3.5$, indicating stable learning across all random seeds. 
Among the compared methods, TD3-NN achieves the highest final reward (around $3.8$), exceeding the GNN-based variants. 
This suggests that, under this operating setting, the policy can be learned effectively from a fixed-dimensional state representation without explicitly encoding network connectivity.
The 69-bus system exhibits a markedly different training profile. 
Rewards start from around $-100$ and end with larger deviations across seeds, ranging from $-10$ to $10$. 
Notably, TD3-GCN and TD3-TAGConv achieve higher final rewards than TD3-NN on average, whereas TD3-GATConv shows strong seed sensitivity and does not consistently outperform the other variants. 
In particular, TD3-GATConv fails to converge under one seed (Fig.~\ref{fig:conver}(b)), indicating that attention-based message passing can offer high representational flexibility but may also introduce optimization instability when the learning problem becomes more challenging. 
For both systems, reward improvements beyond $500$ episodes are marginal, implying diminishing returns from additional training under the current hyperparameter setting.

\begin{figure}[htbp]
    \centering
    \includegraphics[width=1.0\columnwidth]{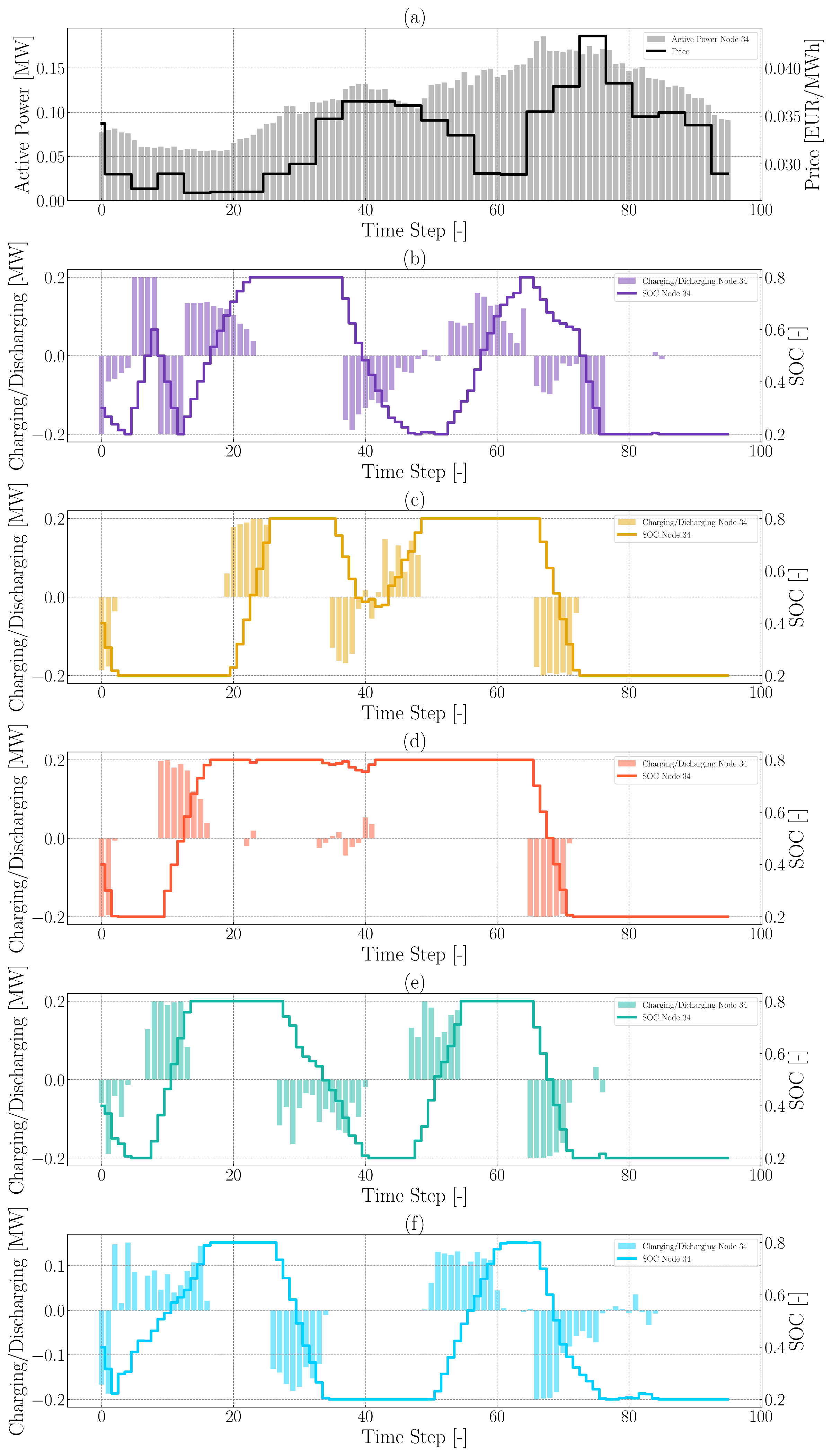}
    \caption{Comparison of RL algorithms on 34 buses system, (a)electricity price and ESS nodes' active power, (b)-(f) ESS nodes' SOC and charging/discharging strategy of NLP, TD3-NN, TD3-GCN, TD3-TAGConv, TD3-GATConv on Node 34.}
    \label{fig:soc_node}
\end{figure}
\begin{figure}[htbp]
    \centering
    \includegraphics[width=1.0\columnwidth]{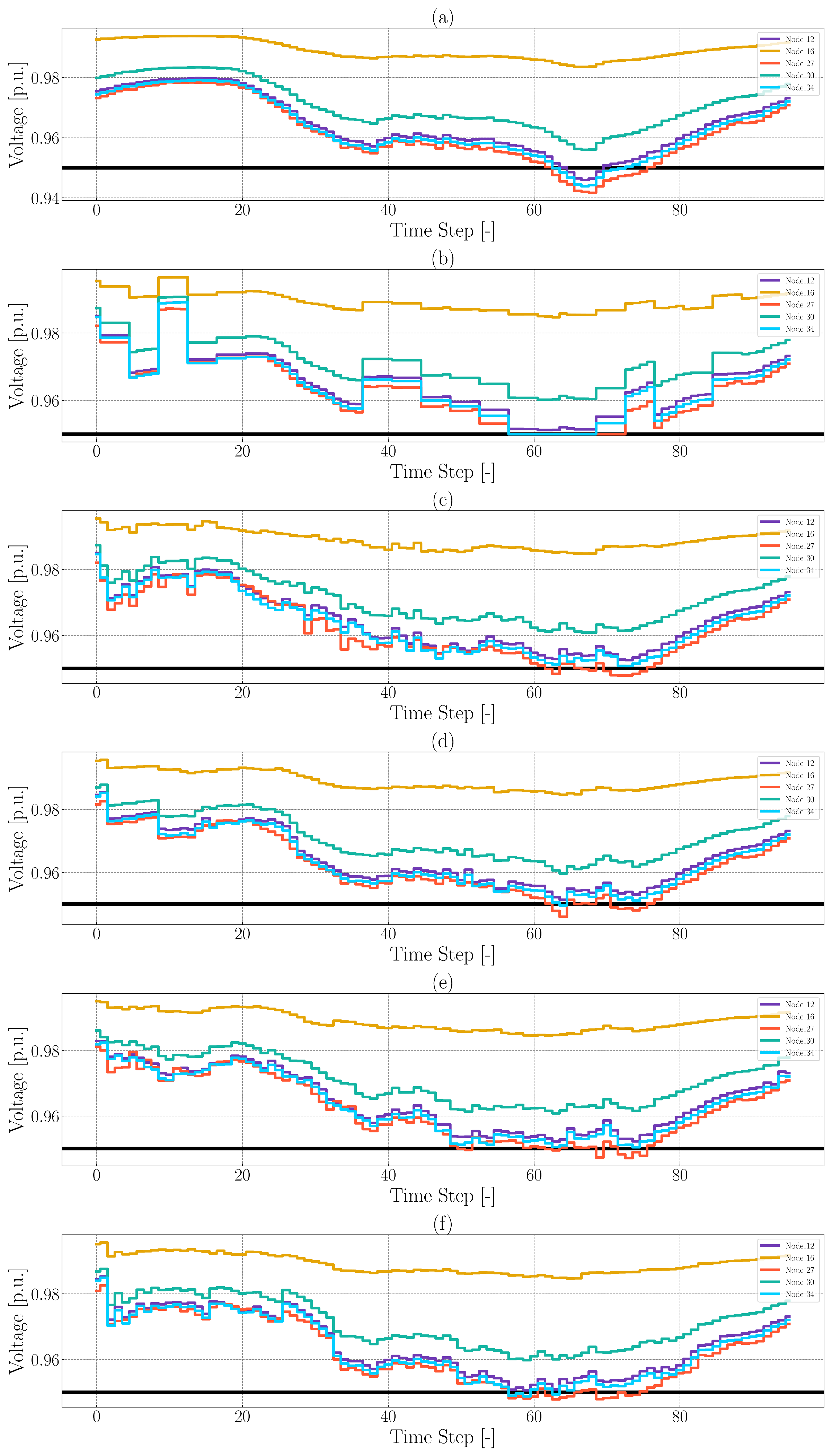}
    \caption{ESS nodes' voltage magnitude on 34 buses system, (a)without ESS control, (b)NLP, (c)TD3-NN, (d)TD3-GCN, (e)TD3-TAGConv, (f)TD3-GATConv.}
    \label{fig:voltage}
\end{figure}
To assess whether the learned policies reproduce meaningful dispatch actions beyond rewards, Fig.~\ref{fig:soc_node} shows the ESS charging/discharging actions and the corresponding SOC trajectories for a representative one-day operation. 
Given the price and net-load profiles in Fig.~\ref{fig:soc_node}(a), the NLP baseline (Fig.~\ref{fig:soc_node}(b)) exhibits two dominant SOC peaks around time steps $23$ and $64$, which serve as reference patterns for evaluating the learned policies.
A closer inspection reveals clear differences during the day. 
For example, at around time step $72$, the NLP solution discharges at the maximum power $200$~kW, reducing the SOC from $0.6$ to $0.2$; this action is economically important for mitigating the cost impact of peak-price periods and operationally beneficial for relieving network stress. 
Most RL algorithms exhibit comparable discharge actions around this period and achieve SOC trajectories that closely track the NLP reference. 
In contrast, TD3-GCN shows a weaker or delayed discharge during time steps $35$--$50$, resulting in a visible SOC mismatch, which is consistent with its lower reward in Fig.~\ref{fig:conver}. 
This indicates that TD3-GCN does not consistently learn the key time-dependent dispatch pattern, particularly the discharge action required at specific critical intervals, potentially because its aggregation tends to smooth node information and provides a less discriminative representation when the optimal action depends on subtle state differences.

Fig.~\ref{fig:voltage} further compares the voltage magnitude trajectories under different dispatch policies. 
Overall, RL-based dispatch mitigates voltage magnitude violations compared with the un-optimized operation, as reflected by fewer violated time steps.
Although some violations remain, this results indicate that the learned policies provide operationally meaningful voltage support in addition to economic cost reduction.

\begin{table*}
\centering
\caption{RL algorithms' average training time [$h/m/s$], mean and $95\%$ confidence bounds of execution time [$s$] and performances on 34 buses testsets.}
\resizebox{\textwidth}{!}{
%\resizebox{\linewidth}{!}{
\begin{tabular}{l|llll|l}
\toprule
\textbf{Algorithms}
& \textbf{TD3-NN}
& \textbf{TD3-GCN}
& \textbf{TD3-TAGConv}
& \textbf{TD3-GATConv}
& \textbf{NLP}\\
\midrule
\textbf{Training [$h/m/s$]}
& \makecell[l]{2h36m6.6s}
& \makecell[l]{5h22m14.7s}
& \makecell[l]{5h58m55.8s}
& \makecell[l]{7h28m51.3s}
& -\\[2mm]
\textbf{Execution [$s$]}
& \makecell[l]{0.0998\\ (0.0990, 0.1007)} 
& \makecell[l]{0.1968\\ (0.1935, 0.2001)}
& \makecell[l]{0.1909\\ (0.1898, 0.1920)}
& \makecell[l]{0.2274\\ (0.2256, 0.2291)}
& \makecell[l]{19.77}\\[4mm]
\makecell[l]{\textbf{Saved costs [\$]}\\}
& \makecell[l]{197.92\\ (195.71, 200.13)}
& \makecell[l]{179.74\\ (175.12, 184.38)}
& \makecell[l]{188.04\\ 183.53, 192.54)} 
& \makecell[l]{195.28\\ (191.48, 199.07)}
& \makecell[l]{260.40\\ (252.34, 268.47)}\\[4mm]
\makecell[l]{\textbf{Accuracy vs NLP [$\%$]}\\}
& \makecell[l]{76.31\\ (75.46, 77.15)}
& \makecell[l]{69.51\\ (67.72, 71.31)}
& \makecell[l]{72.61\\ (70.95, 74.26)} 
& \makecell[l]{75.46\\ (74.02, 76.90)}
& - \\[4mm]
\makecell[l]{\textbf{Accuracy vs TD3-NN [$\%$]}\\}
& -
& \makecell[l]{91.92\\ (89.87, 93.96)}
& \makecell[l]{96\\ (93.58, 98.43)} 
& \makecell[l]{99.57\\ (97.68, 101.46)}
& - \\[4mm]
% \makecell[l]{\textbf{Voltage Violation}\\\textbf{Counts [-]}}
\textbf{Average Voltage Violation Counts [-]}
& 4
& 5
& 4
& 5
& -\\[2mm]
% \makecell[l]{\textbf{Average Voltage}\\\textbf{Violation Value [p.u.]}}
\makecell[l]{\textbf{Average Violation Value [p.u.]}\\}
& \makecell[l]{0.02527\\ (0.02429, 0.02626)} 
& \makecell[l]{0.02492\\ (0.02405, 0.02577)} 
& \makecell[l]{0.02299\\ (0.02212, 0.02385)} 
& \makecell[l]{0.02374\\ (0.02236, 0.02512)} 
& -\\
\bottomrule
\end{tabular}}
\label{tbl:baseline_performance_34}
\end{table*}
\normalsize
%%%%%%%%%%%%%%%%
\begin{table*}
\centering
\caption{RL algorithms' average training time [$h/m/s$], mean and $95\%$ confidence bounds of execution time [$s$] and performances on 69 buses testsets.}
\resizebox{\textwidth}{!}{
%\resizebox{\linewidth}{!}{
\begin{tabular}{l|llll|l}
\toprule
\textbf{Algorithms}
& \textbf{TD3-NN}
& \textbf{TD3-GCN}
& \textbf{TD3-TAGConv}
& \textbf{TD3-GATConv}
& \textbf{NLP}\\
\midrule
\textbf{Training [$h/m/s$]}
& \makecell[l]{2h47m42.6s}
& \makecell[l]{5h32m22.5s}
& \makecell[l]{6h11m58.2s}
& \makecell[l]{7h50m39.6s}
& -\\[2mm]
\textbf{Execution [$s$]}
& \makecell[l]{0.1340\\ (0.1327, 0.1354)} 
& \makecell[l]{0.2351\\ (0.2323, 0.2378)}
& \makecell[l]{0.2381\\ (0.2347, 0.2415)}
& \makecell[l]{0.2878\\ (0.2830, 0.2927)}
& \makecell[l]{144.37}\\[4mm]
\makecell[l]{\textbf{Saved costs [\$]}\\}
& \makecell[l]{458.98\\ (450.91, 467.06)}
& \makecell[l]{498.83\\ (489.58, 508.07)}
& \makecell[l]{466.44\\ (453.93, 478.96)} 
& \makecell[l]{439.42\\ (388.57, 490.281)}
& \makecell[l]{868.19\\ (830.15, 906.25)}\\[4mm]
\makecell[l]{\textbf{Accuracy vs NLP [$\%$]}\\}
& \makecell[l]{54.56\\ (53.69, 55.44)}
& \makecell[l]{59.30\\ (58.21, 60.38)}
& \makecell[l]{55.54\\ (54.25, 56.82)} 
& \makecell[l]{52.74\\ (47.19, 58.29)}
& - \\[4mm]
\makecell[l]{\textbf{Accuracy vs TD3-NN [$\%$]}\\}
& -
& \makecell[l]{109.87\\ (106.65, 113.10)}
& \makecell[l]{102.54\\ (100.86, 104.21)} 
& \makecell[l]{97.99\\ (87.13, 108.86)}
& - \\[4mm]
% \makecell[l]{\textbf{Voltage Violation}\\\textbf{Counts [-]}}
\textbf{Average Voltage Violation Counts [-]}
& 77
& 54
& 33
& 151
& -\\[2mm]
% \makecell[l]{\textbf{Average Voltage}\\\textbf{Violation Value [p.u.]}}
\makecell[l]{\textbf{Average Violation Value [p.u.]}\\}
& \makecell[l]{0.05256\\ (0.05203, 0.05311)} 
& \makecell[l]{0.04786\\ (0.04642, 0.04932)} 
& \makecell[l]{0.04657\\ (0.04454, 0.04861)} 
& \makecell[l]{0.04584\\ (0.04311, 0.04856)} 
& -\\
\bottomrule
\end{tabular}}
\label{tbl:baseline_performance_69}
\end{table*}
\normalsize
Tables~\ref{tbl:baseline_performance_34} and \ref{tbl:baseline_performance_69} report the average training time and execution time (mean and 95\% confidence bounds). 
During training, GNN-based agents require additional computation due to message passing and a larger parameter set; this overhead is expected and mainly affects offline training. 
In contrast, at execution time, the trained policy provides near-instant decisions. 
Compared with solving the NLP online, RL reduces execution time by $86$--$614$$\times$ across all cases, with a larger speedup on the 69-bus system (maximum $100$$\times$ on the 34-bus system and $614$$\times$ on the 69-bus system). 
This highlights a key practical advantage: as network scale increases, optimization-based dispatch becomes more time-consuming, whereas policy inference remains lightweight.
We further decompose performance into economic performance and voltage magnitude constraint satisfaction. 
On the 34-bus system, the saved cost of TD3-NN is $76.31\%$, while TD3-GCN, TD3-TAGConv, and TD3-GATConv achieve $69.51\%$, $72.61\%$, and $99.57\%$, respectively. 
Normalized by the TD3-NN performance, TD3-GCN, TD3-TAGConv, and TD3-GATConv reach $91.92\%$, $96.00\%$, and $99.57\%$ of the TD3-NN saved cost, respectively, showing comparable economic performance to the NN baseline.
On the 69-bus system, TD3-GCN and TD3-TAGConv surpass TD3-NN ($54.56\%$) with mean accuracies of $59.3\%$ and $55.54\%$, respectively, while TD3-GATConv yields a slightly lower mean accuracy of $52.74\%$ but a higher 95\% upper bound ($58.29\%$) than TD3-NN. 
This discrepancy is mainly driven by one non-convergent seed in TD3-GATConv, which lowers the mean while preserving strong best-case performance.
For voltage magnitude regulation, GNN-based methods consistently achieve fewer voltage violations than TD3-NN, indicating that topology-aware representations improve constraint-related decision-making. 
A comparison among GNN variants also reveals different trade-offs. 
On the 69-bus system, TD3-GCN attains higher economic accuracy but shows more voltage violations than TD3-TAGConv, suggesting that it tends to prioritize reward and economic objectives while sacrificing constraint satisfaction. 
In contrast, TD3-TAGConv and TD3-GATConv better balance economic gains and voltage constraints, likely because their broader (multi-hop) or adaptive (attention-based) aggregation captures medium-range couplings that are critical for voltage magnitude regulation in deeper radial structures.

The overall results indicate that learning difficulty depends strongly on network size and topology. 
The 69-bus system, with longer branches, deeper radial depth, and more heterogeneous coupling, induces a larger state space and more complex action propagation, leading to higher reward variance and lower average accuracy than the 34-bus system. 
In such settings, GNN-based policies show clearer advantages over TD3-NN, because message passing explicitly encodes electrical connectivity and helps the policy infer how local ESS actions influence network-wide voltages and flows.

Among the GNN variants, TD3-TAGConv and TD3-GATConv perform strongly on the 34-bus system, which can be attributed to their higher representational capacity. 
TAGConv effectively enlarges the receptive field via multi-hop aggregation, capturing dependencies beyond immediate neighbors, while GATConv uses attention to emphasize electrically influential neighbors and suppress less relevant signals. 
On the 69-bus system, TD3-GATConv achieves a higher upper bound but a lower mean than TD3-TAGConv, highlighting a capacity--stability trade-off: higher expressiveness can improve best-case performance but may increase sensitivity to random seeds. 
Overall, these results suggest that topology-aware learning is beneficial for larger and structurally more complex distribution networks, while further stabilization is still needed for attention-based architectures.

\subsection{Generalization to Re-configured Topologies}
To assess generalization beyond the training topology, we evaluate the learned policies under topology changes. 
We first test transfer to reconfigured topologies within the same network, where the bus set is unchanged and only branch connections are modified. 
We then consider a more challenging setting by transferring policies across different benchmark systems (34-bus and 69-bus), where both network size and structure differ.
\subsubsection{In-network reconfiguration}
Table~\ref{tbl:topo_reconfig_34} and Table~\ref{tbl:topo_reconfig_69} summarize the topology reconfiguration cases for the 34-bus and 69-bus systems, respectively.
The reconfiguration set covers three levels of structural change: 
(i) two cases that modify one connection within a branch, 
(ii) two cases that modify two connections within a branch, and 
(iii) two cases that reconnect an entire branch to a different location.
The specific node and branch changes for each topology are listed in the corresponding tables.
\begin{table}[htbp]
\centering
\caption{Topological Variation Scenarios or Evaluating the Generalization of GNN-based RL Algorithms on 34 Buses System.}
\resizebox{\columnwidth}{!}{
\begin{tabular}{lll}
\toprule
\textbf{ID} &  \textbf{Reconfiguration} & \textbf{Connections} \\
\midrule
\textbf{TP1}  & --     & Baseline topology \\
\midrule
\textbf{TP2}  & 1 within a branch      & (25,26) $\Rightarrow$ (24,26)  \\
\textbf{TP3}  & 1 within a branch      &  (32,33) $\Rightarrow$ (31,33)\\
\textbf{TP4}  & 2 within a branch      &  (11,12) $\Rightarrow$ (10,12); (29,30) $\Rightarrow$ (28,30)\\
\textbf{TP5}  & 2 within a branch     &  (15,16) $\Rightarrow$ (14,16); (33,34) $\Rightarrow$ (32,34)\\
\textbf{TP6}  & 1 whole branch     & (10,31) $\Rightarrow$ (8,31) \\
\textbf{TP7}  & 1 whole branch     &  (10,11) $\Rightarrow$ (9,11)\\
\bottomrule
\end{tabular}}
\label{tbl:topo_reconfig_34}
\end{table}
\begin{table}[htbp]
\centering
\caption{Topological Variation Scenarios or Evaluating the Generalization of GNN-based RL Algorithms on 69 Buses System.}
\resizebox{\columnwidth}{!}{
\begin{tabular}{lll}
\toprule
\textbf{ID} &  \textbf{Reconfiguration} & \textbf{Connections} \\
\midrule
\textbf{TP1}  & --     & Baseline topology \\
\midrule
\textbf{TP2}  & 1 within a branch      & (67,68) $\Rightarrow$ (13,68)  \\
\textbf{TP3}  & 1 within a branch      & (45,46) $\Rightarrow$ (44,46)  \\
\textbf{TP4}  & 2 within a branch      & (35,36) $\Rightarrow$ (34,36); (13,69) $\Rightarrow$ (14,69)  \\
\textbf{TP5}  & 2 within a branch     & (50,51) $\Rightarrow$ (49,51); (52,53) $\Rightarrow$ (9,53) \\
\textbf{TP6}  & 1 whole branch     & (12,13) $\Rightarrow$ (11,13)\\
\textbf{TP7}  & 1 whole branch     &  (10,54) $\Rightarrow$ (8,54)\\
\bottomrule
\end{tabular}}
\label{tbl:topo_reconfig_69}
\end{table}

\begin{figure*}[htbp]
    \centering
    \includegraphics[width=2.0\columnwidth]{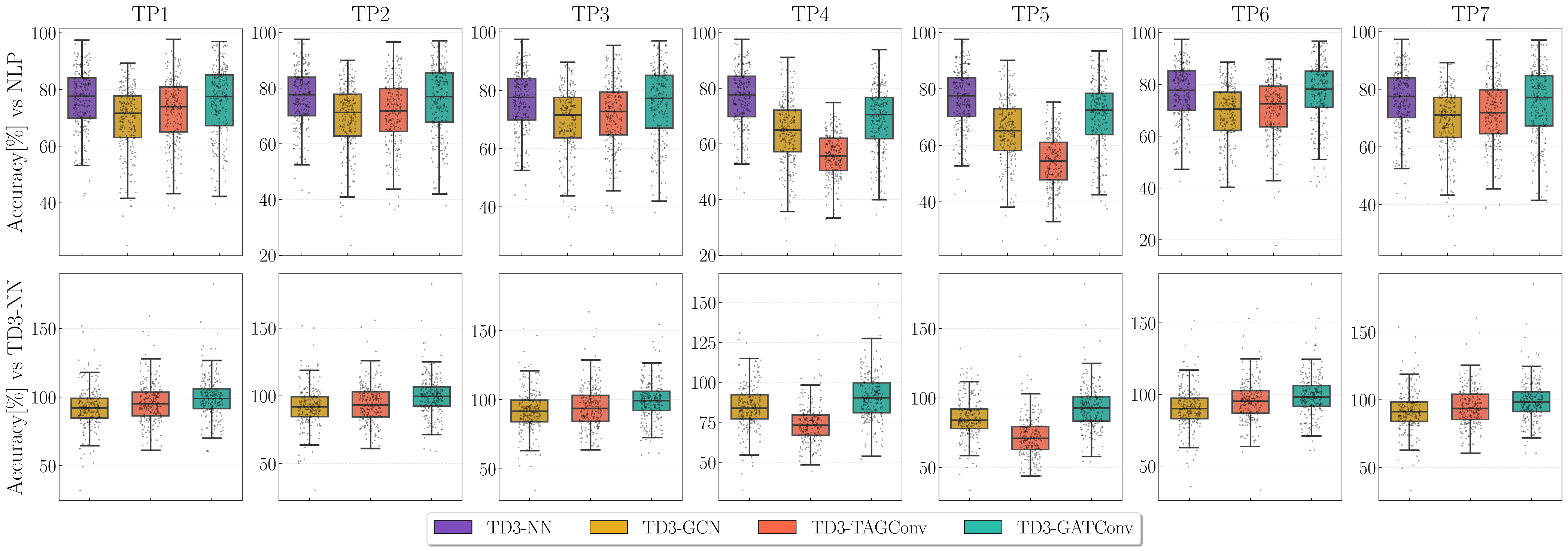}
    \caption{34 buses system RL algorithms' operational accuracy [$\%$] on topological variations}
    \label{fig:violin_accuracy_34}
\end{figure*}

\begin{figure*}[htbp]
    \centering
    \includegraphics[width=2.0\columnwidth]{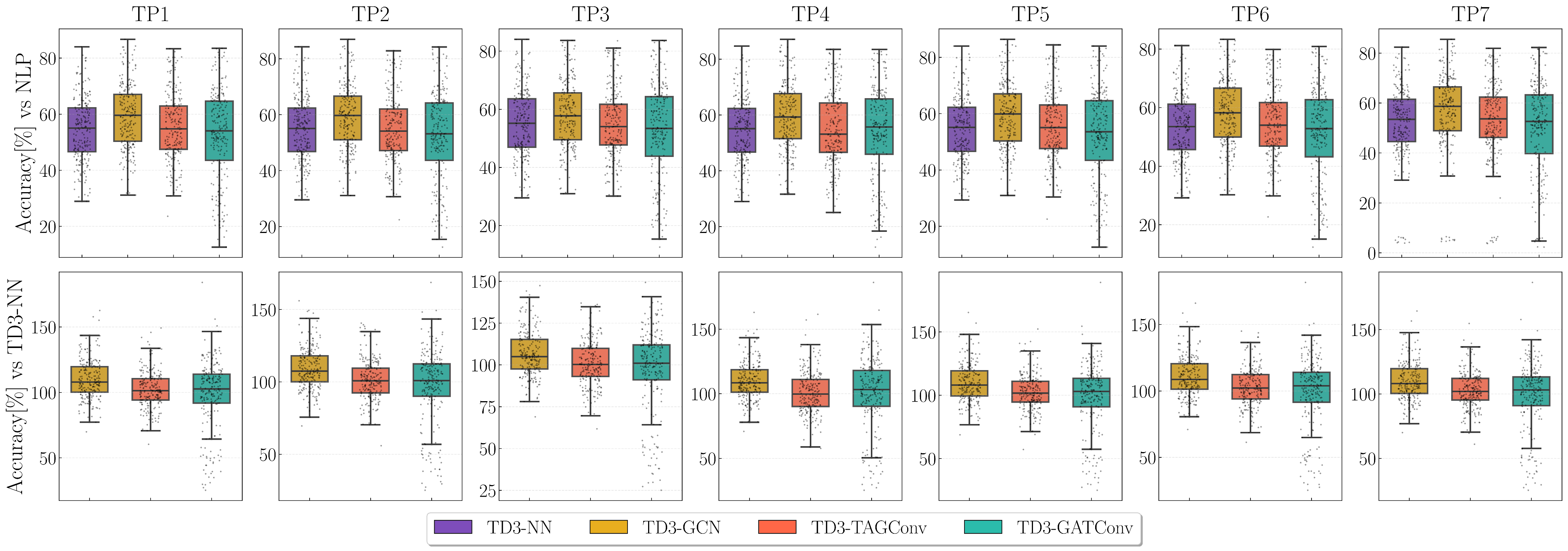}
    \caption{69 buses system RL algorithms' operational accuracy [$\%$] on topological variations}
    \label{fig:violin_accuracy_69}
\end{figure*}

Fig.~\ref{fig:violin_accuracy_34} and Fig.~\ref{fig:violin_accuracy_69} report the saved cost accuracy across seven topologies for the two systems, including both the comparison of RL methods against the NLP baseline and the comparison between GNN-based and NN-based controllers.
Overall, policies trained on the original topology can be deployed on reconfigured networks with only modest performance degradation.
Across most topology cases, the average accuracy remains close to the original-topology performance (34-bus: $75\%$; 69-bus: $55\%$), indicating that the learned dispatch strategy is not overly dependent on a single fixed topology and remains effective under moderate structural perturbations.
The results also reveal that transfer is feasible for both NN-based and GNN-based controllers because the bus set is unchanged under reconfiguration, allowing the same state construction to be applied.
However, their behavior differs in a way that is consistent with the topology-awareness of the model.
Specifically, the NN-based controller can operate without explicitly encoding connectivity, but its performance and constraint handling may vary when reconfiguration changes how ESS actions propagate through the network.
In contrast, the GNN-based controllers incorporate the updated adjacency through message passing, which helps maintain more consistent performance across reconfigured cases, particularly when the structural change alters electrical coupling relationships.

The relative ranking among GNN variants is largely preserved under reconfiguration.
On the 34-bus system, the average accuracy continues to follow TD3-GCN $\lesssim$ TD3-TAGConv $\lesssim$ TD3-GATConv across most cases, while on the 69-bus system it follows TD3-GCN $\gtrsim$ TD3-TAGConv $\gtrsim$ TD3-GATConv (see Fig.~\ref{fig:violin_accuracy_34}, \ref{fig:violin_accuracy_69}).

This consistency suggests that the architectural differences observed on the training topology generally persist under reconfiguration.
Nevertheless, notable exceptions indicate architecture-specific sensitivity to certain perturbations.
For the 34-bus system, TD3-TAGConv shows a larger accuracy drop to $55.71\%$ and $54.34\%$ in TP4 and TP5, which correspond to within-branch reconfigurations with two connection changes.
A plausible explanation is that TAGConv’s $k$-hop aggregation depends on the stability of multi-hop neighborhoods; when two connections are altered within a branch, the $k$-hop receptive field can change substantially, leading to a mismatch between the learned neighborhood patterns and the new graph structure.
This sensitivity is less pronounced for TD3-GATConv, whose attention mechanism can adaptively reweight neighbors and reduce the impact of newly introduced, less informative messages after reconfiguration.
On the 69-bus system, TP4 shows that TD3-GATConv achieves higher mean accuracy than TD3-TAGConv, even though this ordering does not hold on the training topology.
This result highlights the transfer potential of attention-based aggregation under certain structural shifts.
At the same time, the mean performance of TD3-GATConv is affected by training instability under one random seed (as discussed in Fig.~\ref{fig:conver}), suggesting that improved stabilization is important to fully realize its generalization advantage.

\begin{figure}[htbp]
    \centering
    \includegraphics[width=1.0\columnwidth]{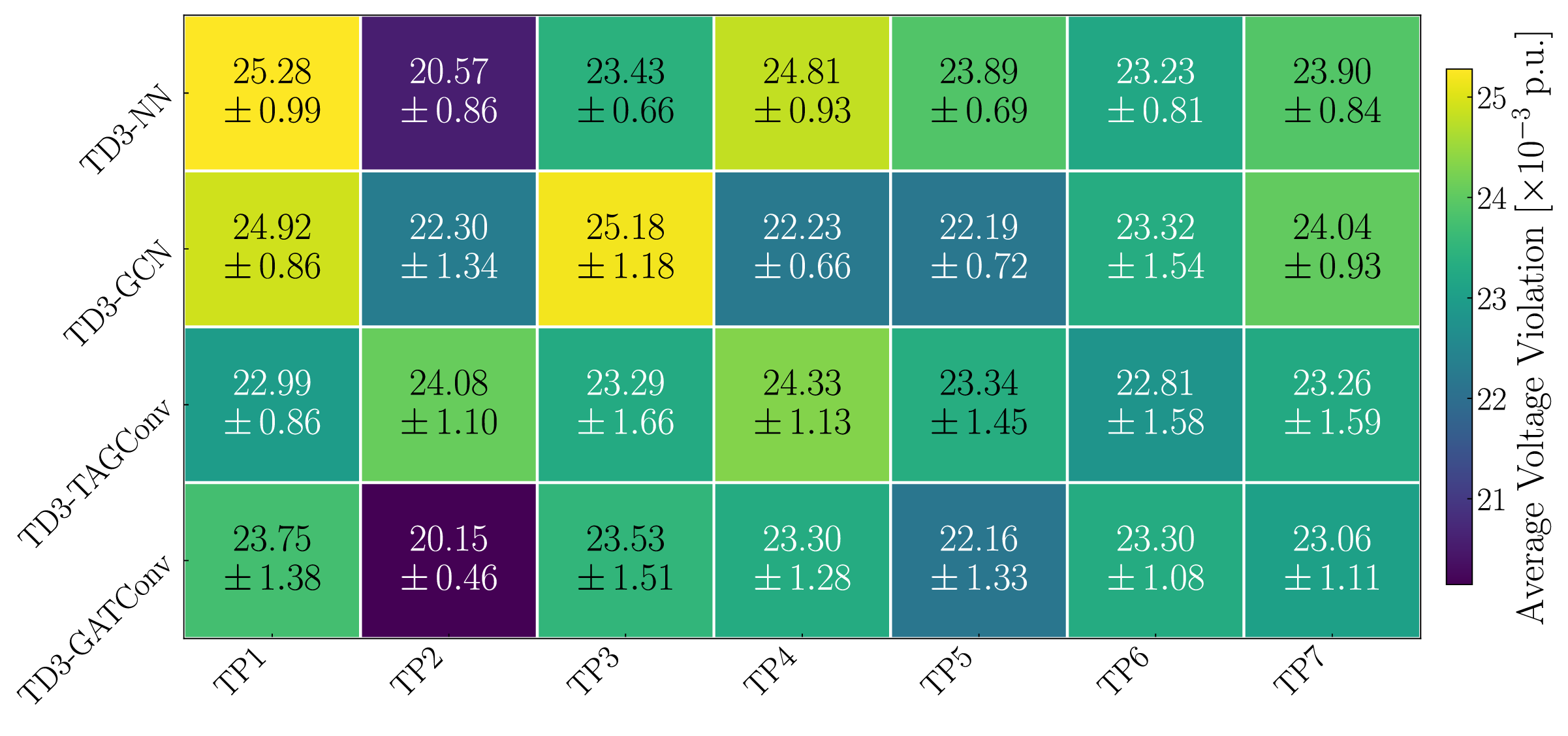}
    \caption{34 buses system RL algorithms' mean $95\%$ confidence bounds of average voltage violation [p.u.] on topological variations.}
    \label{fig:bubble_voltage_34}
\end{figure}
\begin{figure}[htbp]
    \centering
    \includegraphics[width=1.0\columnwidth]{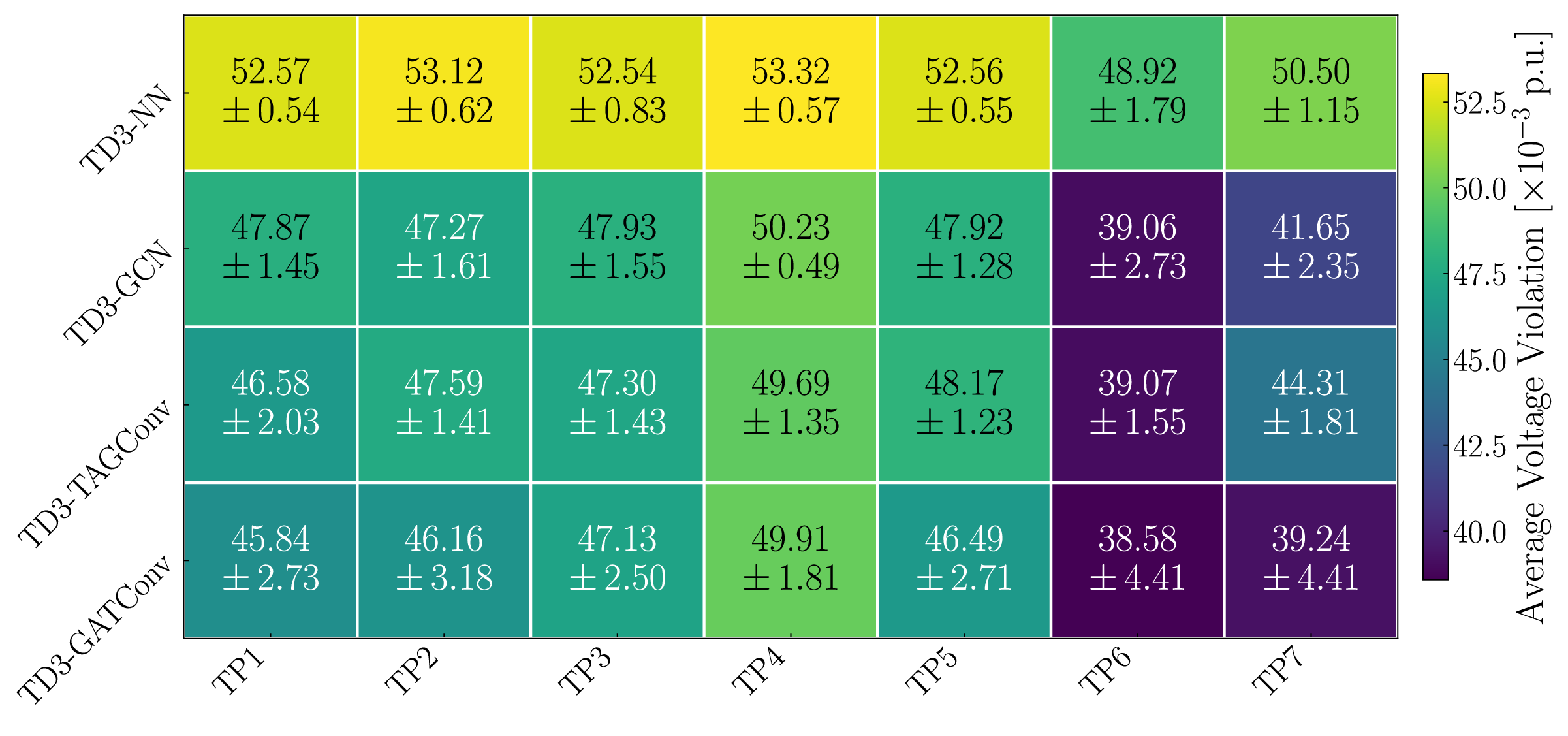}
    \caption{69 buses system RL algorithms' mean $95\%$ confidence bounds of average voltage violation [p.u.] on topological variations.}
    \label{fig:bubble_voltage_69}
\end{figure}
Fig.~\ref{fig:bubble_voltage_34} and Fig.~\ref{fig:bubble_voltage_69} report the mean voltage-violation metrics with 95\% confidence intervals under reconfigured topologies.
On the 34-bus system, GNN-based controllers achieve fewer voltage violations than TD3-NN in most cases, except TP2 and TP3 where the advantage becomes less evident.
On the 69-bus system, all GNN-based controllers consistently yield lower voltage violations than the NN-based controller across all reconfiguration cases.
These findings suggest that the benefit of topology-aware representations becomes more pronounced as system complexity increases.
In larger networks with deeper radial structures and longer electrical coupling paths, local ESS actions can influence voltage profiles in a more topology-dependent way.
GNN message passing explicitly captures such connectivity-mediated propagation, which helps the controller anticipate constraint-related impacts under new topologies, thereby improving voltage regulation robustness.

\subsubsection{Cross-network transfer}
\begin{figure}[htbp]
    \centering
    \includegraphics[width=1.0\columnwidth]{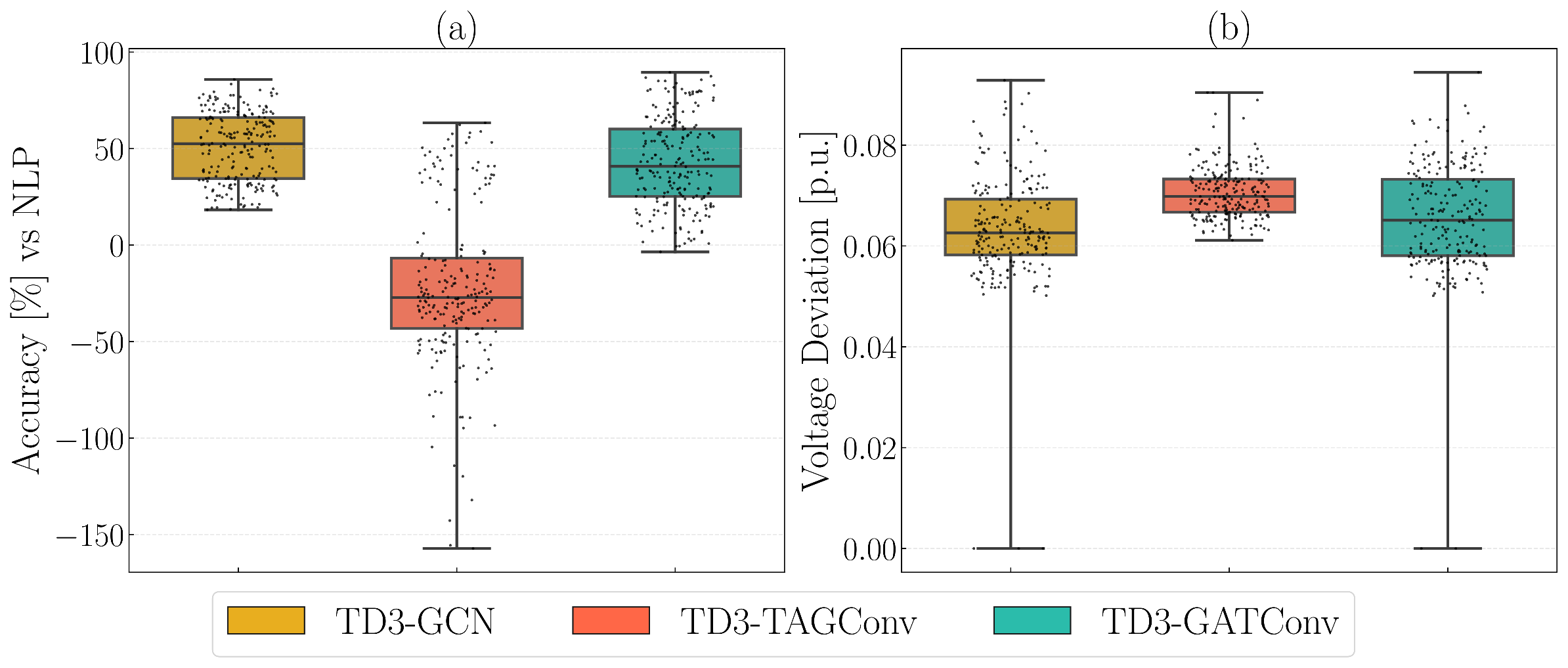}
    \caption{RL algorithms' (a)operational accuracy [$\%$] and (b) mean $95\%$ confidence bounds of average voltage violation [p.u.] on 69 buses system with agents trained on 34 buses system}
    \label{fig:34_69}
\end{figure}

\begin{figure}[htbp]
    \centering
    \includegraphics[width=1.0\columnwidth]{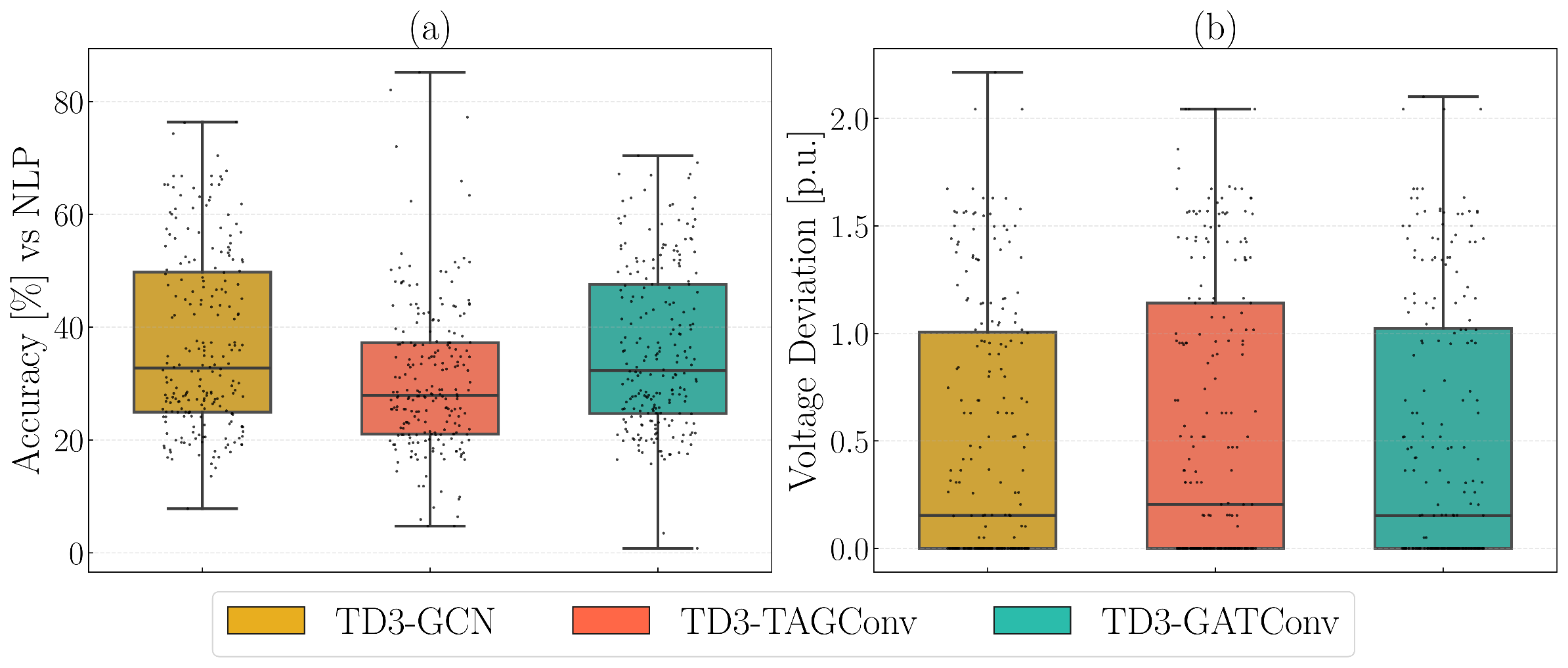}
    \caption{RL algorithms' (a)operational accuracy [$\%$] and (b) mean $95\%$ confidence bounds of average voltage violation [p.u.] on 34 buses system with agents trained on 69 buses system}
    \label{fig:69_34}
\end{figure}

We further evaluate transfer to a completely different topology with a different number of buses by applying models trained on one system to the other.
The results are shown in Fig.~\ref{fig:34_69} (34$\rightarrow$69) and Fig.~\ref{fig:69_34} (69$\rightarrow$34).
Because the NN-based controller requires a fixed input dimension tied to the bus set, only GNN-based controllers can be directly transferred across systems with different node counts under the current state definition.

Despite this compatibility, cross-system transfer leads to a substantial performance drop.
For 69$\rightarrow$34, the saved cost accuracy decreases to approximately $50\%$, and TD3-TAGConv even results in negative savings (i.e., higher energy cost than the pre-optimization baseline), indicating a clear failure mode under severe distribution shift.
For 34$\rightarrow$69, the accuracy drops further to around $30\%$, reflecting the difficulty of transferring a policy learned on a small system to a significantly larger and structurally different network.
In both directions, voltage violations increase compared with within-system reconfiguration transfer, suggesting that constraint-relevant knowledge does not generalize reliably under large structural shifts.

These results support two key conclusions.
First, transferring to a fundamentally different topology is unlikely to maintain the same accuracy level, because the learned dispatch pattern and constraint-response relationship can become invalid when the underlying power flow coupling structure changes substantially.
Second, the asymmetric transfer performance (69$\rightarrow$34 outperforming 34$\rightarrow$69) implies that policies learned on a more complex system capture richer and more generalizable representations of network interactions, whereas policies learned on a simpler network may lack the capacity to extrapolate to the more complex coupling patterns in a larger system.
Nevertheless, the remaining gap indicates that direct zero-shot transfer is insufficient; more systematic adaptation, such as fine-tuning on the target system or training with topology randomization, is needed to achieve reliable cross-system generalization.

\section{Conclusion}\label{sec:conclusion}

This paper develops and evaluates topology-aware GNN-based reinforcement learning for real-time, constraint-aware ESS dispatch in distribution networks. 
We implement a GNN-enhanced actor-critic RL framework with multiple graph encoders (GCN, TAGConv, and GATConv) and benchmark it against NN and global-optimization baselines on 34-bus and 69-bus systems, including tests under topology reconfiguration and cross-system transfer.

The results show that all RL controllers enable faster online dispatch, reducing execution time by $86$--$614\times$ compared with solving the NLP. 
GNN-based policies consistently achieve fewer time steps with voltage magnitude violations and lower violation magnitudes, and this advantage is more evident in the larger, more topologically complex 69-bus system and under topology reconfiguration. 
In terms of economic performance, the NN baseline is competitive on the 34-bus system, while on the 69-bus system TD3-GCN and TD3-TAGConv achieve on average $109.32\%$ and $101.67\%$ of the TD3-NN performance across training and reconfigured topology, respectively.
Moreover, zero-shot transfer to a fundamentally different graph remains challenging in this task, as economic performance drops substantially and voltage violations increase under severe structural shift; GAT-based encoders may achieve strong best-case transfer but can be sensitive to training instability.
Future work will focus on improving robustness through topology-diversified training and lightweight adaptation on the target network.

\appendix
% \section{ADD HERE IF YOU NEED AN APPENDIX}

\printcredits
%% Loading bibliography style file
% \bibliographystyle{model1-num-names}
% \bibliographystyle{cas-model2-names}
\bibliographystyle{IEEEtran}
% Loading bibliography database
\bibliography{cas-refs}

@article{wachter2006implementation,
  title={On the implementation of an interior-point filter line-search algorithm for large-scale nonlinear programming},
  author={W{\"a}chter, Andreas and Biegler, Lorenz T},
  journal={Mathematical programming},
  volume={106},
  number={1},
  pages={25--57},
  year={2006},
  publisher={Springer}
}

@article{solheim2024visualizing,
  title={Visualizing graph neural networks in order to learn general concepts in power systems},
  author={Solheim, {\O}ystein Rognes and Presthus, Gunnhild Svandal and H{\o}verstad, Boye Annfelt and Korp{\aa}s, Magnus},
  journal={Electric Power Systems Research},
  volume={237},
  pages={110717},
  year={2024},
  publisher={Elsevier}
}

@article{wu2023two,
  title={Two-stage voltage regulation in power distribution system using graph convolutional network-based deep reinforcement learning in real time},
  author={Wu, Huayi and Xu, Zhao and Wang, Minghao and Zhao, Jian and Xu, Xu},
  journal={International Journal of Electrical Power \& Energy Systems},
  volume={151},
  pages={109158},
  year={2023},
  publisher={Elsevier}
}

@article{gao2025graph,
  title={Graph reinforcement learning for real-time dynamic reconfiguration and fault management in energy storage networks},
  author={Gao, Yingqi and Zhang, Zhanqiang and Meng, Keqilao and Liu, Wenyu and Gao, Ruifeng},
  journal={Journal of Energy Storage},
  volume={125},
  pages={116833},
  year={2025},
  publisher={Elsevier}
}

@article{chen2023physical,
  title={Physical-assisted multi-agent graph reinforcement learning enabled fast voltage regulation for PV-rich active distribution network},
  author={Chen, Yongdong and Liu, Youbo and Zhao, Junbo and Qiu, Gao and Yin, Hang and Li, Zhengbo},
  journal={Applied Energy},
  volume={351},
  pages={121743},
  year={2023},
  publisher={Elsevier}
}

@article{xiang2023deep,
  title={Deep reinforcement learning based topology-aware voltage regulation of distribution networks with distributed energy storage},
  author={Xiang, Yue and Lu, Yu and Liu, Junyong},
  journal={Applied Energy},
  volume={332},
  pages={120510},
  year={2023},
  publisher={Elsevier}
}

@article{xu2024optimal,
  title={An optimal solutions-guided deep reinforcement learning approach for online energy storage control},
  author={Xu, Gaoyuan and Shi, Jian and Wu, Jiaman and Lu, Chenbei and Wu, Chenye and Wang, Dan and Han, Zhu},
  journal={Applied Energy},
  volume={361},
  pages={122915},
  year={2024},
  publisher={Elsevier}
}

@article{henry2021gym,
  title={Gym-ANM: Reinforcement learning environments for active network management tasks in electricity distribution systems},
  author={Henry, Robin and Ernst, Damien},
  journal={Energy and AI},
  volume={5},
  pages={100092},
  year={2021},
  publisher={Elsevier}
}

@article{hossain2025topology,
  title={Topology-aware reinforcement learning for voltage control: Centralized and decentralized strategies},
  author={Hossain, Rakib and Gautam, Mukesh and MansourLakouraj, Mohammad and Livani, Hanif and Benidris, Mohammed},
  journal={IEEE Transactions on Industry Applications},
  year={2025},
  publisher={IEEE}
}

@article{donon2024topology,
  title={Topology-aware reinforcement learning for tertiary voltage control},
  author={Donon, Balthazar and Cubelier, Francois and Karangelos, Efthymios and Wehenkel, Louis and Crochepierre, Laure and Pache, Camille and Saludjian, Lucas and Panciatici, Patrick},
  journal={Electric Power Systems Research},
  volume={234},
  pages={110658},
  year={2024},
  publisher={Elsevier}
}

@article{lin2024powerflownet,
  title={PowerFlowNet: Power flow approximation using message passing Graph Neural Networks},
  author={Lin, Nan and Orfanoudakis, Stavros and Cardenas, Nathan Ordonez and Giraldo, Juan S and Vergara, Pedro P},
  journal={International Journal of Electrical Power \& Energy Systems},
  volume={160},
  pages={110112},
  year={2024},
  publisher={Elsevier}
}

@article{licari2025addressing,
  title={Addressing voltage regulation challenges in low voltage distribution networks with high renewable energy and electrical vehicles: A critical review},
  author={Licari, John and Rhaili, Salah Eddine and Micallef, Alexander and Staines, Cyril Spiteri},
  journal={Energy Reports},
  volume={14},
  pages={2977--2997},
  year={2025},
  publisher={Elsevier}
}

@article{abu2018modern,
  title={Modern network reconfiguration techniques for service restoration in distribution systems: A step to a smarter grid},
  author={Abu-Elanien, Ahmed EB and Salama, MMA and Shaban, Khaled B},
  journal={Alexandria engineering journal},
  volume={57},
  number={4},
  pages={3959--3967},
  year={2018},
  publisher={Elsevier}
}

@article{hou2024distflow,
  title={Distflow safe reinforcement learning algorithm for voltage magnitude regulation in distribution networks},
  author={Hou, Shengren and Fu, Aihui and Duque, Edgar Mauricio Salazar and Palensky, Peter and Chen, Qixin and Vergara, Pedro P},
  journal={Journal of Modern Power Systems and Clean Energy},
  volume={13},
  number={1},
  pages={300--311},
  year={2024},
  publisher={SGEPRI}
}

@inproceedings{gao2025symbolic,
  title={Symbolic Deep Reinforcement Learning for Energy Storage Systems Optimal Dispatch},
  author={Gao, Shuyi and Hou, Shengren and Palensky, Peter and Vergara, Pedro P},
  booktitle={2025 IEEE Kiel PowerTech},
  pages={1--5},
  year={2025},
  organization={IEEE}
}

@article{deng2025dr,
  title={DR-RQL: A Sustainable Demand Response-Based Learning System for Energy Scheduling and Battery Health Estimation},
  author={Deng, Kailian and Zhang, Hongtao and Cui, Zihao and Zha, Zhongyi and Gao, Shuyi and Yan, Shuai and Hua, Yicun and Liu, Xiaojie and Xu, Shaoxuan and Wei, Fang and others},
  journal={Sustainability},
  volume={17},
  number={24},
  pages={10970},
  year={2025},
  publisher={MDPI}
}

@inproceedings{gao2024linear,
  title={Linear reinforcement learning for energy storage systems optimal dispatch},
  author={Gao, Shuyi and Hou, Shengren and Duque, Edgar Mauricio Salazar and Palensky, Peter and Vergara, Pedro P},
  booktitle={2024 IEEE PES Innovative Smart Grid Technologies Europe (ISGT EUROPE)},
  pages={1--6},
  year={2024},
  organization={IEEE}
}

@article{vergara2020stochastic,
  title={A stochastic programming model for the optimal operation of unbalanced three-phase islanded microgrids},
  author={Vergara, Pedro P and L{\'o}pez, Juan Camilo and Rider, Marcos J and Shaker, Hamid R and da Silva, Luiz CP and J{\o}rgensen, Bo N},
  journal={International Journal of Electrical Power \& Energy Systems},
  volume={115},
  pages={105446},
  year={2020},
  publisher={Elsevier}
}

@article{xiong2024two,
  title={Two-Stage Robust Optimal Operation of Distribution Networks Considering Renewable Energy and Demand Asymmetric Uncertainties},
  author={Xiong, Zhisheng and Zeng, Bo and Palensky, Peter and Vergara, Pedro P},
  journal={arXiv preprint arXiv:2411.10166},
  year={2024}
}

@article{li2024optimal,
  title={Optimal Power Flow in a highly renewable power system based on attention neural networks},
  author={Li, Chen and Kies, Alexander and Zhou, Kai and Schlott, Markus and El Sayed, Omar and Bilousova, Mariia and St{\"o}cker, Horst},
  journal={Applied Energy},
  volume={359},
  pages={122779},
  year={2024},
  publisher={Elsevier}
}

@article{jaradat2025review,
  title={A review of battery energy storage system for renewable energy penetration in electrical power system: environmental impact, sizing methods, market features, and policy frameworks},
  author={Jaradat, Tha'er and Khatib, Tamer},
  journal={Future Batteries},
  pages={100106},
  year={2025},
  publisher={Elsevier}
}

@article{shengren2023optimal,
  title={Optimal energy system scheduling using a constraint-aware reinforcement learning algorithm},
  author={Shengren, Hou and Vergara, Pedro P and Duque, Edgar Mauricio Salazar and Palensky, Peter},
  journal={International Journal of Electrical Power \& Energy Systems},
  volume={152},
  pages={109230},
  year={2023},
  publisher={Elsevier}
}

@article{xing2023real,
  title={Real-time optimal scheduling for active distribution networks: A graph reinforcement learning method},
  author={Xing, Qiang and Chen, Zhong and Zhang, Tian and Li, Xu and Sun, KeHui},
  journal={International Journal of Electrical Power \& Energy Systems},
  volume={145},
  pages={108637},
  year={2023},
  publisher={Elsevier}
}

@inproceedings{rossi2022unreasonable,
  title={On the unreasonable effectiveness of feature propagation in learning on graphs with missing node features},
  author={Rossi, Emanuele and Kenlay, Henry and Gorinova, Maria I and Chamberlain, Benjamin Paul and Dong, Xiaowen and Bronstein, Michael M},
  booktitle={Learning on graphs conference},
  pages={11--1},
  year={2022},
  organization={PMLR}
}

@article{azarnia2024offering,
  title={Offering of active distribution network in real-time energy market by integrated energy management system and Volt-Var optimization},
  author={Azarnia, Mahsa and Rahimiyan, Morteza and Siano, Pierluigi},
  journal={Applied Energy},
  volume={358},
  pages={122635},
  year={2024},
  publisher={Elsevier}
}

@misc{DHPC2024,
author = {{D}elft {H}igh {P}erformance {C}omputing {C}entre ({DHPC})},
title = {{D}elft{B}lue {S}upercomputer ({P}hase 2)},
year = {2024},
howpublished = {\url{https://www.tudelft.nl/dhpc/ark:/44463/DelftBluePhase2}},
ark = {ark:/44463/DelftBluePhase2}
}

@article{velivckovic2017graph,
  title={Graph attention networks},
  author={Veli{\v{c}}kovi{\'c}, Petar and Cucurull, Guillem and Casanova, Arantxa and Romero, Adriana and Lio, Pietro and Bengio, Yoshua},
  journal={arXiv preprint arXiv:1710.10903},
  year={2017}
}

@article{brody2021attentive,
  title={How attentive are graph attention networks?},
  author={Brody, Shaked and Alon, Uri and Yahav, Eran},
  journal={arXiv preprint arXiv:2105.14491},
  year={2021}
}

@article{du2017topology,
  title={Topology adaptive graph convolutional networks},
  author={Du, Jian and Zhang, Shanghang and Wu, Guanhang and Moura, Jos{\'e} MF and Kar, Soummya},
  journal={arXiv preprint arXiv:1710.10370},
  year={2017}
}

@article{kipf2016semi,
  title={Semi-supervised classification with graph convolutional networks},
  author={Kipf, TN},
  journal={arXiv preprint arXiv:1609.02907},
  year={2016}
}

@article{lillicrap2015continuous,
  title={Continuous control with deep reinforcement learning},
  author={Lillicrap, Timothy P and Hunt, Jonathan J and Pritzel, Alexander and Heess, Nicolas and Erez, Tom and Tassa, Yuval and Silver, David and Wierstra, Daan},
  journal={arXiv preprint arXiv:1509.02971},
  year={2015}
}

@inproceedings{fujimoto2018addressing,
  title={Addressing function approximation error in actor-critic methods},
  author={Fujimoto, Scott and Hoof, Herke and Meger, David},
  booktitle={International conference on machine learning},
  pages={1587--1596},
  year={2018},
  organization={PMLR}
}

@ARTICLE{9906438,
  author={Xing, Qiang and Xu, Yan and Chen, Zhong and Zhang, Ziqi and Shi, Zhao},
  journal={IEEE Transactions on Industrial Informatics}, 
  title={A Graph Reinforcement Learning-Based Decision-Making Platform for Real-Time Charging Navigation of Urban Electric Vehicles}, 
  year={2023},
  volume={19},
  number={3},
  pages={3284-3295},
  doi={10.1109/TII.2022.3210264}}

@article{hou2024rl,
  title={RL-ADN: A high-performance deep reinforcement learning environment for optimal energy storage systems dispatch in active distribution networks},
  author={Hou, Shengren and Gao, Shuyi and Xia, Weijie and Duque, Edgar Mauricio Salazar and Palensky, Peter and Vergara, Pedro P},
  journal={Energy and AI},
  volume={19},
  pages={100457},
  year={2025},
  publisher={Elsevier}
}

@article{garcia2021stochastic,
  title={Stochastic optimization of microgrids with hybrid energy storage systems for grid flexibility services considering energy forecast uncertainties},
  author={Garcia-Torres, Felix and Bordons, Carlos and Tobajas, Javier and Real-Calvo, Rafael and Santiago, Isabel and Grieu, Stephane},
  journal={IEEE Trans. on Power Systems},
  volume={36},
  number={6},
  pages={5537--5547},
  year={2021},
  publisher={IEEE}
}

@article{lu2021multistage,
  title={Multistage robust optimization of routing and scheduling of mobile energy storage in coupled transportation and power distribution networks},
  author={Lu, Zhuoxin and Xu, Xiaoyuan and Yan, Zheng and Shahidehpour, Mohammad},
  journal={IEEE Trans. on Transportation Electrification},
  volume={8},
  number={2},
  pages={2583--2594},
  year={2021},
  publisher={IEEE}
}

@article{Rev_MINLP2,
	author = {Miros{\l}aw Parol and Tomasz W{\'o}jtowicz and Krzysztof Ksi{\k e}{\.z}yk and Christoph Wenge and Stephan Balischewski and Bartlomiej Arendarski},
	doi = {https://doi.org/10.1016/j.ijepes.2020.105886},
	issn = {0142-0615},
	journal = {Int. J. of Electrical Power \& Energy Systems},
	pages = {105886},
	title = {Optimum management of power and energy in low voltage microgrids using evolutionary algorithms and energy storage},
	volume = {119},
	year = {2020}}

@article{Rev_3,
	author = {Zhang, Zidong and Zhang, Dongxia and Qiu, Robert C.},
	doi = {10.17775/CSEEJPES.2019.00920},
	journal = {CSEE J. of Power and Energy Systems},
	number = {1},
	pages = {213-225},
	title = {Deep reinforcement learning for power system applications: An overview},
	volume = {6},
	year = {2020}}

\end{document}